\documentclass[10pt,twocolumn,letterpaper]{article}

\usepackage{cvpr}              
\usepackage{pifont}
\usepackage{float}
\usepackage{stfloats}
\usepackage{arydshln}
[p]
\definecolor{cvprblue}{rgb}{0.21,0.49,0.74}
\usepackage[pagebackref,breaklinks,colorlinks,allcolors=cvprblue]{hyperref}
\usepackage{multirow}
\def\paperID{16791} 
\def\confName{CVPR}
\def\confYear{2025}

\title{Paint by Inpaint: Learning to Add Image 
Objects by Removing Them First}
\newcommand{\pipe}{\textsc{PIPE}}
\definecolor{inpaintBlue}{HTML}{9AC2E3}
\definecolor{paintOrange}{HTML}{F09C8F}

\author{
    Navve Wasserman\textsuperscript{1*}, \ Noam Rotstein\textsuperscript{2*}, \ Roy Ganz\textsuperscript{2}, \ Ron Kimmel\textsuperscript{2}   \\
    {\normalsize \textsuperscript{1} Weizmann Institute of Science \quad \textsuperscript{2} Technion - Israel Institute of Technology} \\
    \tt\small{*Indicates equal contribution.}
}

\begin{document}
\twocolumn[{
\maketitle
\vspace{-0.78cm} 
\begin{center}
    \captionsetup{type=figure}
    \includegraphics[width=0.88\textwidth]{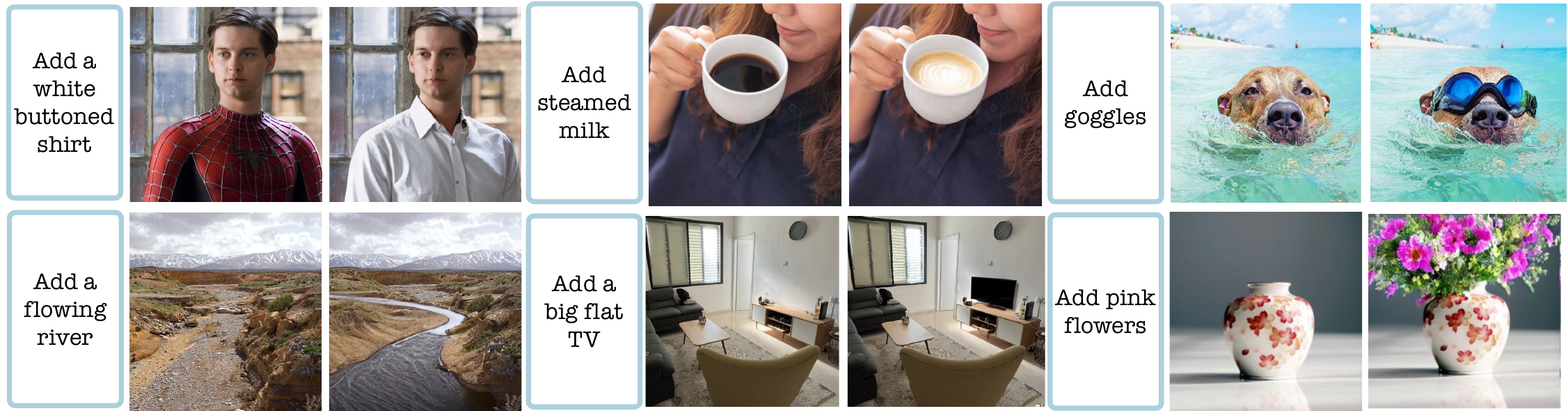}
    \vspace{-0.22cm} 
        \caption{\textbf{Visual Results of the Proposed Models.}}
          \label{Fig:Visual_Result_Teaser}
    \vspace{-0.05cm} 
\end{center}
}]

\begin{abstract}
Image editing has advanced significantly with the introduction of text-conditioned diffusion models. Despite this progress, seamlessly adding objects to images based on textual instructions without requiring user-provided input masks remains a challenge. 
We address this by leveraging the insight that removing objects (\texttt{Inpaint}) is significantly simpler than its inverse process of adding them (\texttt{Paint}),
attributed to inpainting models that benefit from segmentation mask guidance.
Capitalizing on this realization, by implementing an automated and extensive pipeline, we curate a filtered large-scale image dataset containing pairs of images and their corresponding object-removed versions.
Using these pairs, we train a diffusion model to inverse the inpainting process, effectively adding objects into images.
Unlike other editing datasets, ours features natural target images instead of synthetic ones while ensuring source-target consistency by construction.
Additionally, we utilize a large Vision-Language Model to provide detailed descriptions of the removed objects and a Large Language Model to convert these descriptions into diverse, natural-language instructions.
Our quantitative and qualitative results show that the trained model surpasses existing models in both object addition and general editing tasks.
Visit our \href{https://rotsteinnoam.github.io/Paint-by-Inpaint}{project page} for the released dataset and trained models.
\end{abstract}

    
\vspace{-0.5cm}
\section{Introduction}\label{sec:intro}
\vspace{-0.2cm}
Image editing plays a central role in the computer vision and graphics communities, with diverse applications spanning various domains.
The task is inherently challenging as each image offers infinite editing possibilities, each with countless potential outcomes.
A particularly intricate editing task is seamlessly adding objects to images, which requires not only realistic visuals but also a nuanced understanding of the global image context, including parameters such as location, scale, and style.
While many solutions require the user to provide a mask for the target object \citep{li2023gligen,xie2023smartbrush,rombach2022high,wang2023imagen}, recent advancements have capitalized on the success of text-conditioned diffusion models to enable a mask-free approach \citep{brooks2023instructpix2pix, zhang2023hive}.
Such solutions offer a more convenient and realistic setting; yet, they still encounter challenges, as demonstrated in \Cref{Fig:Visual_Comparison}.

The leading method for such editing, InstructPix2Pix (IP2P) \citep{brooks2023instructpix2pix}, synthesizes a dataset containing triplets of source and target images alongside an editing instruction as guidance.
Under this guidance, a model is trained to transform source images into target ones.
While showing some success, the model's effectiveness is bounded by the quality of the synthesized training data.
We address this limitation by introducing an alternative automatic method for creating a large-scale, high-quality dataset for image object addition.
Our approach is grounded in the observation that adding objects (\texttt{paint}) is essentially the inverse of removing them (\texttt{inpaint}).
Namely, by using pairs of images—ones containing objects and others with objects removed—an object addition dataset can be established.
In practice, we create the dataset by leveraging abundant images and object masks available in segmentation datasets~\citep{kuznetsova2020open, lin2014microsoft, gupta2019lvis} alongside a high-end inpainting model~\citep{rombach2022high}.
The outputs are then used in a reverse manner, with the original images as editing targets and the inpainted ones as sources.
This reversed approach is essential because directly adding objects with an inpainting model requires object segmentations not present in the images.
\textbf{
Our approach offers two key advantages over IP2P:
}
(i)
While IP2P relies on synthetic source and target images, our targets are real natural images, with source images also being natural outside the typically small edited regions.
(ii)
Despite employing techniques such as prompt-to-prompt~\citep{hertz2022prompt} and Directional CLIP-based filtering~\citep{gal2021stylegan} to address source-target consistency issues, IP2P often fails to achieve this.
Our approach inherently maintains consistency by construction.

Mask-based inpainting models have recently shown great success in filling image masks naturally and coherently~\citep{rombach2022high}.
However, since these models were not trained specifically for object removal, their use for this purpose is not guaranteed to be artifact-free,
potentially leaving remnants of the original object, unintentionally creating new objects, or causing other distortions.
Given that the outputs of inpainting serve as training data, these artifacts could potentially impair the performance of the resulting models.
To counteract these issues, we propose a comprehensive pipeline of varied filtering and refinement techniques.
Additionally, we complement the source and target image pairs with natural language editing instructions by harnessing advancements in multimodal learning~\citep{li2023blip,dai2023instructblip,liu2023improved,bai2023qwen,ganz2023models,ganz2024question, rotstein2023fusecap}.
By employing a Large Vision-Language Model (VLM)~\citep{wang2024cogvlm}, we generate elaborated captions for the target objects.
Next, we utilize a Large Language Model (LLM)~\citep{jiang2023mistral} to cast these descriptions to natural language instructions for object addition.
To further enhance our dataset, we incorporate human-annotated object reference datasets~\citep{kazemzadeh2014referitgame, mao2016generation} and convert them into adding instructions.
Overall, we combine these sources to form an instruction-based object addition dataset, named \pipe{} (\textbf{P}aint by \textbf{I}n\textbf{p}aint \textbf{E}diting). 
Unprecedented in size, our dataset features approximately $1$ million image pairs, spans over $1400$ different classes, and includes thousands of unique attributes.

Utilizing \pipe{}, we train a diffusion model to follow object addition instructions, setting a new standard for adding realistic image objects, as demonstrated in \Cref{Fig:Visual_Result_Teaser}, and as validated across extensive experiments on multiple benchmarks.
Besides quantitative results, we conduct a human evaluation survey comparing our model to top-performing models, showcasing its improved capabilities.
Furthermore, we demonstrate that \pipe{} can extend beyond mere object addition; by integrating it with additional editing datasets, we show it significantly improves overall editing results.


\noindent
\textbf{Our contributions include:} 
\begin{itemize}
\item Introduction of the {\it Paint by Inpaint} framework for image editing.
\item Construction of \pipe{}, a large-scale, high-quality, mask-free, textual instruction-guided object addition image dataset.
\item Demonstration of a diffusion-based model trained with \pipe{},  achieving state-of-the-art object addition to images and enhanced general editing performance.

\end{itemize}

\vspace{-0.3cm}
\begin{figure*}[t]
    \centering
      \includegraphics[width=0.9\linewidth]{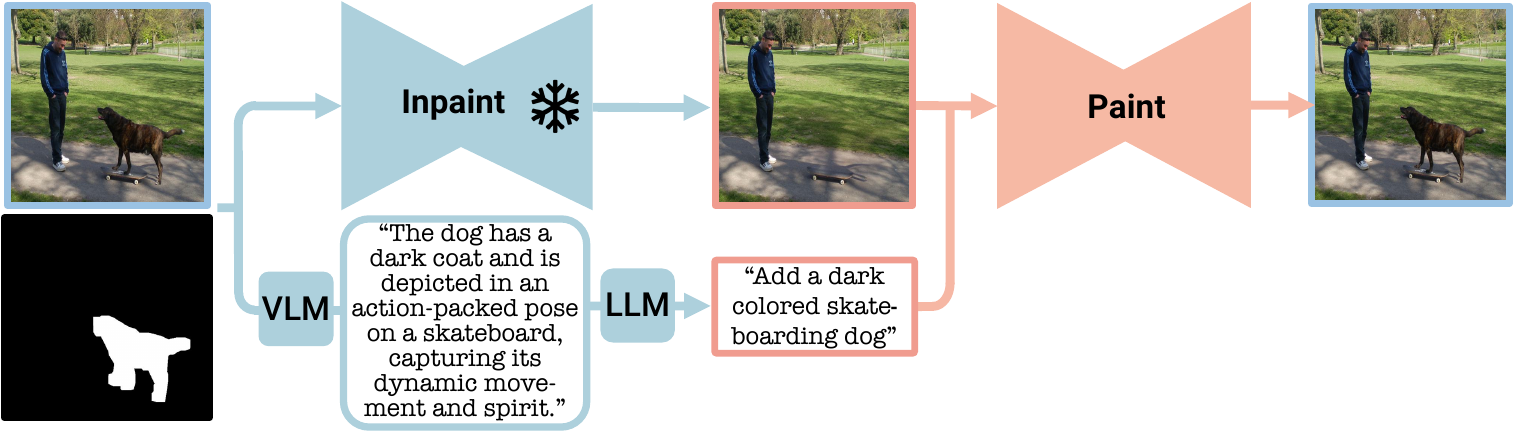}
    \vspace{-0.1cm}

    \caption{
    \textbf{Paint by Inpaint Framework.}
    Illustration of our two-phase approach: 
    (1) Building \pipe{} dataset (\textcolor{inpaintBlue}{blue}), which involves:
    (i) Removing the object utilizing a frozen inpainting model and the object mask.
    (ii) Generating addition instructions, demonstrated through the VLM-LLM-based procedure, where a VLM extracts visual object details and an LLM formulates them into instructions.
    (2) Training an editing model (\textcolor{paintOrange}{orange}), \pipe{} is employed to train a model to reverse the inpainting process, thereby adding objects to images.
    }
    \vspace{-0.35cm}
    \label{method}
\end{figure*}

\vspace{0.05cm}
\section{Related Efforts}\label{sec:related}
\vspace{-0.1cm}
\subsection{Image Editing}
\vspace{-0.12cm}
Image editing has long been explored in computer graphics and vision~\citep{oh2001image, perez2023poisson}.
The field has seen substantial advances with the emergence of diffusion-based image synthesis models~\citep{song2020score, ho2020denoising}, especially with their text-conditioned variants~\citep{ramesh2022hierarchical, rombach2022high, saharia2022photorealistic, nichol2021glide}.
The application of such models can be broadly categorized into two distinct approaches -- mask-based and mask-free.

\textbf{Mask-Based Editing. }
Such approaches formulate image editing as an inpainting task, using a mask to outline the target edit region.
Early diffusion-based techniques utilized pretrained models for inpainting~\citep{song2020score, avrahami2022blended, yu2023inpaint, meng2021sdedit}, while more recent approaches fine-tune the models specifically for this task~\citep{nichol2021glide, saharia2022palette, rombach2022high}.
Inpainting models benefit from the possibility of training on large-scale image datasets, as they can be trained with any image paired with a random mask.
Various attempts have been made to advance this methodology in different directions~\citep{wang2023imagen, li2023gligen, xie2023smartbrush},
but despite this progress, relying on a user-provided mask makes this setting less preferable in real-world applications.



\textbf{Mask-Free Editing. }
This paradigm allows image editing using text and natural language as an intuitive interactive tool without the need for additional masks.
Kawar \textit{et al.}~\citep{kawar2023imagic} optimize a model to align its output with a target embedding text.
Bar Tal \textit{et al}.~\citep{bar2022text2live} introduces a model that merges an edit layer with the original image.
IP2P turns mask-free image editing into a supervised task by generating an instruction-based dataset using an LLM and Prompt-to-Prompt~\citep{hertz2022prompt}, which adjusts cross-attention layers in diffusion models to align attention maps between source and target prompts.
Rotstein \textit{et al}.~\cite{rotstein2024pathways} take a different approach by leveraging video generators.
These mask-free techniques are distinguished by their ability to perform global edits such as style transfer. 
However, they exhibit limitations in local edits, specifically in maintaining consistency outside the desired edit region.
IP2P seeks to address this by utilizing Directional CLIP loss \citep{gal2021stylegan} for dataset filtering.
Nevertheless, it mitigates the limitation, but only to some extent.
In contrast, our dataset ensures consistency by strictly limiting changes to the intended edit regions only.

\textbf{Instructions-Based Editing. }
A few studies have introduced textual instructions for intuitive, mask-free image editing without complex prompts~\citep{el2019tell, zhang2021text}.
IP2P facilitates this by leveraging GPT-3~\citep{brown2020language} to create editing instructions from input image captions.
Following the advancements in instruction-following capabilities of LLMs~\citep{ouyang2022training, ziegler2019fine}, Zhang \textit{et al.} devise a reward function reflecting user preferences on edited images~\citep{zhang2023hive}.
Our approach takes a different course; it enriches the class-based instructions constructed from the segmentation datasets by employing a VLM~\citep{wang2023cogvlm} to comprehensively describe the target object, and an LLM \citep{jiang2023mistral} to transform the VLM outputs into coherent editing instructions.
Our dataset is further enhanced by integrating object reference datasets \citep{kazemzadeh2014referitgame, mao2016generation}.

\subsection{Image Editing Datasets}\label{related:datasets}
\vspace{-0.1cm}
Early editing approaches~\citep{xu2018attngan, zhang2017stackgan} used datasets with specific classes without direct correspondence between source and target images \citep{lin2014microsoft, wah2011caltech, nilsback2008automated}.
Building datasets of natural images and their natural edited versions
in the mask-free setting is infeasible, as it requires two identical images differing solely in the edited region.
Thus, previous works propose synthetic alternatives, with the previously discussed IP2P's dataset being one of the most prominent ones.
MagicBrush \citep{zhang2024magicbrush} recently introduced a partially synthetic dataset, which was manually created using DALL-E2 \citep{ramesh2022hierarchical}.
While offering more accuracy and consistency, its manual annotation and monitoring limit its scalability.
Inst-Inpaint \citep{yildirim2023inst} leverages segmentation and inpainting models to develop a dataset focused on object removal, designed to eliminate the segmentation step.
We introduce a high-quality image editing dataset that exceeds the scale of any currently available ones.
Furthermore, our approach, uniquely leverages real images as the edit targets, distinguishing it from prior datasets consisting of synthetic data.

\vspace{-0.35cm}
\begin{figure*}[t]
  \centering
\includegraphics[width=0.95\linewidth]{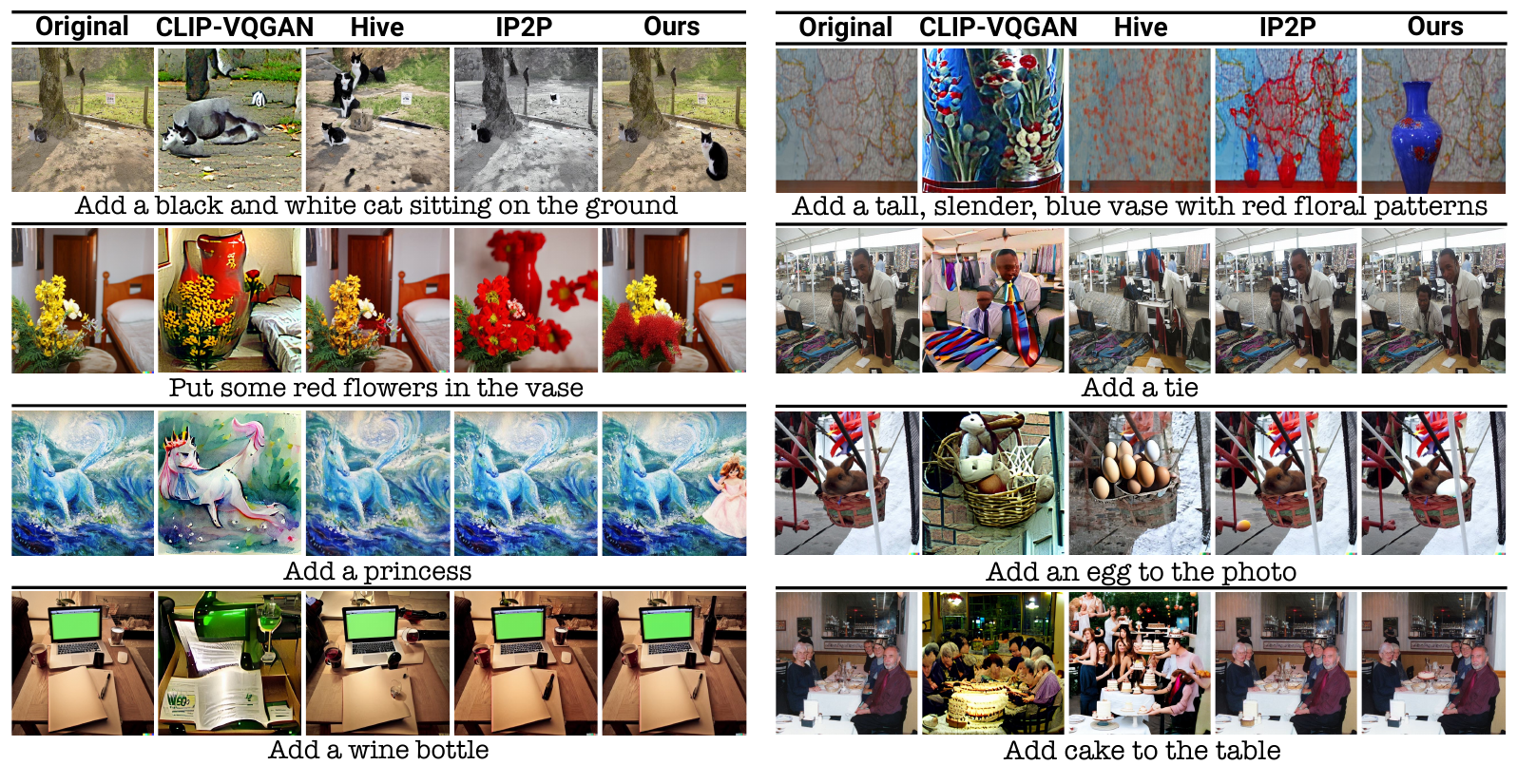}
   \vspace{-0.5cm}
  \caption{\textbf{Visual Comparison.} 
  Comparison of our model with leading editing models across benchmarks, demonstrating superior fidelity to instructions and precise object addition in terms of style, scale, and position, while maintaining higher consistency with original images.
  }
 \vspace{-0.36cm}
  \label{Fig:Visual_Comparison}
\end{figure*}

\vspace{+0.2cm}
\subsection{Object Focused Editing}
\vspace{-0.15cm}
Processing specific objects through diffusion models has gained significant attention in recent research.
For instance, various methodologies have been developed to generate images of particular subjects~\citep{ruiz2023dreambooth, gal2022image, chen2024subject}.
Within the editing domain, Wang \textit{et al.}\citep{wang2023imagen} concentrate on mask-based object editing, training their model for inpainting within existing object boundaries, while Patashnik \textit{et al.}\citep{patashnik2023localizing} introduce a technique for producing diverse variations of such objects.
Similar to our work, SmartBrush~\citep{xie2023smartbrush} aims to add objects to images.
However, unlike our methodology, it requires an input mask from the user.
Instruction-based methods like IP2P and MagicBrush highlight their capability to insert image objects,
allocating a considerable portion of their dataset for this purpose,
for example,
$39\%$ of the MagicBrush dataset is dedicated to this task.
Following the initial release of our paper, several works have emerged with related ideas \cite{canberk2024erasedraw, tarres2024thinking}.

\vspace{-0.2cm}
\section{PIPE Dataset}\label{sec:dataset}
\vspace{-0.2cm}
As outlined in \Cref{sec:related}, leading mask-free, instruction-following image editing models are trained on datasets that are either small-scale or synthetic and inconsistent.
To enhance the efficacy of these models, we propose a systematic method to create a dataset that addresses these limitations.
The devised dataset, dubbed \pipe{} (\textbf{P}aint by \textbf{I}n\textbf{P}aint \textbf{E}dit),
comprises approximately $1$ million image pairs accompanied by diverse object addition instructions.
Our methodology, illustrated in blue in \Cref{method}, unfolds in a two-stage procedure.
First, drawing on the insight that object removal is more straightforward than object addition, we create pairs of source and target images—without and with objects.
Subsequently, we generate a natural language object addition instruction for each pair using various techniques.
In the following section, we describe the proposed pipeline in detail.

\vspace{-0.15cm}
\subsection{Generating Source-Target Image Pairs}\label{sec:method:dataset}
\vspace{-0.1cm}
In the initial stage of creating \pipe{}, we leverage extensive image segmentation datasets.
Specifically, we utilize COCO~\citep{lin2014microsoft} and Open Images~\citep{OpenImages}, enriched with segmentation mask annotations from LVIS~\citep{gupta2019lvis}.
Unifying these datasets results in $889,230$ unique images with over $1,400$ object classes.
We use this diverse corpus for object removal using a Stable Diffusion (SD)~\citep{rombach2022high} based inpainting model\footnote{\url{https://hf.co/runwayml/stable-diffusion-inpainting}}.
This configuration is the underlying reason why constructing \pipe{} via removal is more straightforward than via addition.
However, since the inpainting model was not trained specifically for object removal, it can yield suboptimal outcomes, e.g., leaving original object traces or generating new objects.
To address this, we implement a pipeline of pre-removal and post-removal steps.

\begin{figure*}[t]
    \centering
    \begin{subfigure}[b]{0.17\textwidth}
        \centering
        \includegraphics[width=\textwidth]{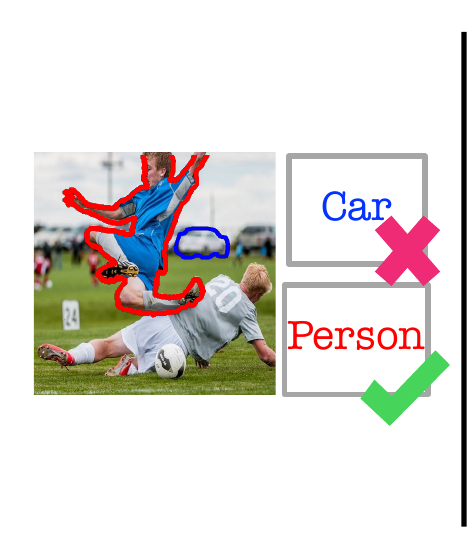}
        \vspace{-0.45cm}
        {\captionsetup{justification=centering}
        \caption{Pre-Removal\\ Abnormal View \label{Fig:filter:a}}
        }
    \end{subfigure}
    \hfill 
    \begin{subfigure}[b]{0.4\textwidth}
        \centering
        \includegraphics[width=\textwidth]{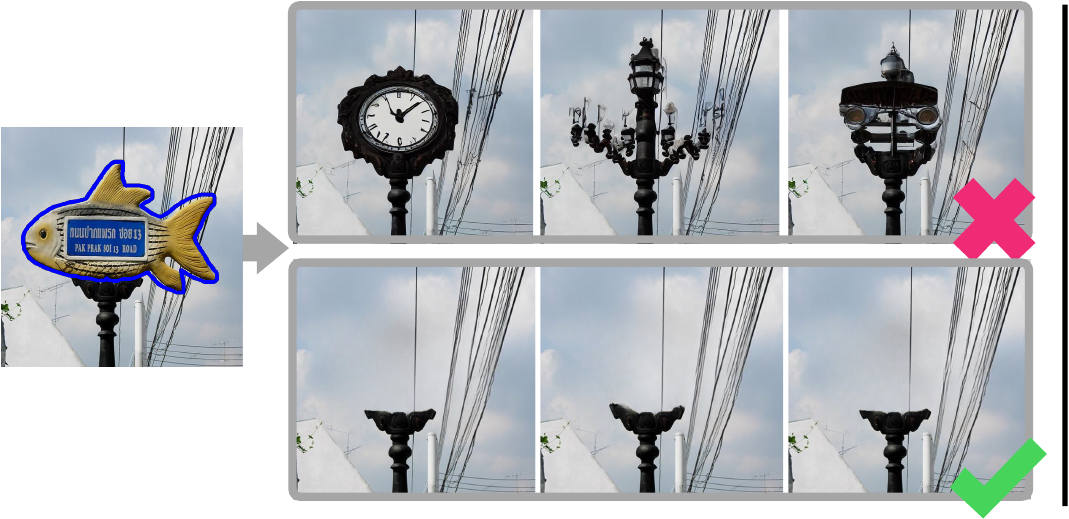}
        \vspace{-0.4cm}
        {
        \captionsetup{justification=centering}
        \caption{Post-Removal\\ CLIP Consensus}
        \label{Fig:filter:b}
        }
    \end{subfigure}
    \hfill 
    \begin{subfigure}[b]{0.3\textwidth}
        \centering
        \includegraphics[width=\textwidth]{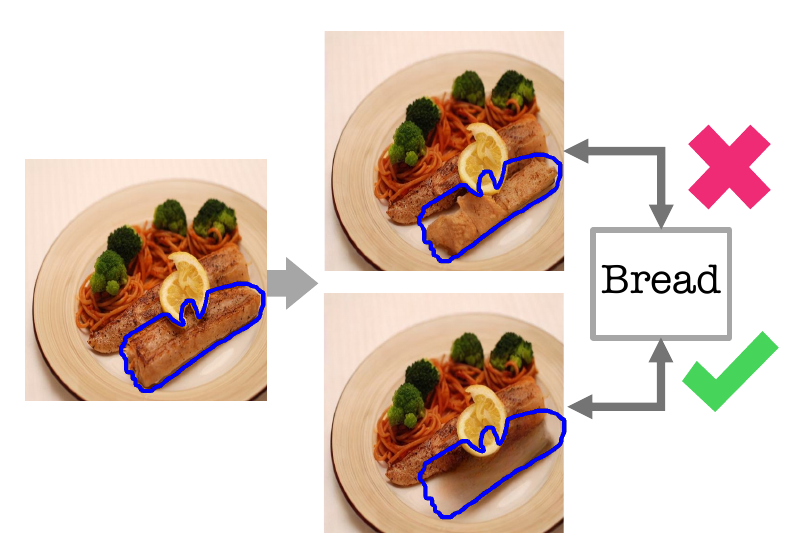}
        \vspace{-0.4cm}
        {
        \captionsetup{justification=centering}
        \caption{Post-Removal\\ Multimodal CLIP}
        \label{Fig:filter:c}
        }
    \end{subfigure}
    \vspace{-0.14cm}
    \caption{\textbf{Dataset Filtering Stages.}
    In constructing \pipe{}, several filtering stages address inpainting drawbacks.
    Initially, a pre-removal filter targets abnormal object views due to blur and low quality.
    Subsequently, a post-removal inconsistency filter identifies a lack of CLIP consensus among three inpainting outputs, indicating substantial variance and potential object regeneration.
    Finally, a post-removal multimodal CLIP filtering ensures low semantic similarity with the original object name.
    }
    \vspace{-0.45cm}
    \label{Fig:filter}
\end{figure*}

\noindent
\textbf{Pre-Removal. }
Object segmentation masks filtering step, retaining only candidates suitable for the subsequent object-adding.
First, we exclude masks according to their size (too large or too small) and location (near image borders).
Next, we use CLIP to calculate the semantic similarity between segmented objects and their class names, using low values to filter out abnormal object views (\textit{e.g.}, blurred objects) and non-informative partial views (\textit{e.g.}, occluded objects).
In \Cref{Fig:filter:a}, we provide an example of a car being filtered due to its small size and blur, while a person without these characteristics is not (see \cref{SM_figure:pre_removal} for more examples).
To ensure the mask fully covers the object, we apply morphological dilation, a crucial step since any unmasked object parts can lead the inpainting model to regenerate it \citep{pobitzer2024outline}.

\vspace{-0.35cm}

\vspace{0.33cm}
\noindent
\textbf{Object Removal. }
Given the dilated masks, we remove the objects using the SD inpainting model.
Unlike conventional inpainting objectives, which aim at general image completion, our focus centers on object removal.
To this end, we guide the model with positive and negative prompts designed to replace objects with non-objects (\textit{e.g.}, background).
The positive prompt is set to ``\texttt{a photo of a background, a photo of an empty place}'',
while the negative prompt is defined as ``\texttt{an object, a <class>}'', where \texttt{<class>} denotes the object class name.
During the inpainting process, we utilize $10$ diffusion steps and generate $3$ distinct outputs per input.

\noindent
\textbf{Post-Removal. }
The last part of our removal pipeline involves employing a multi-step process aimed at filtering and refining the inpainting outputs:
\begin{itemize}[leftmargin=8pt]
    \item \underline{Removal Verification}:
    For each source image and its three inpainted outputs, we introduce two mechanisms to assess removal effectiveness.
    First, we measure the semantic diversity of the three inpainted candidates' regions by calculating the standard deviation of their CLIP embeddings, a metric we refer to as the CLIP consensus.
    Intuitively, high diversity (no consensus) suggests failed object removal, leaving varied non-background object elements, as shown in the upper row of \Cref{Fig:filter:b}.
    Conversely, lower variability (consensus) points to a consistent removal, increasing the likelihood of an appropriate background, as demonstrated in the bottom row of the figure.
    Next, we calculate the CLIP similarity between the inpainted region of each candidate and the class name of the removed object (e.g., \texttt{<bread>}).
     This procedure, referred to as multimodal CLIP filtering, is illustrated in  \Cref{Fig:filter:c}. Introducing CLIP consensus and multimodal CLIP filtering mechanisms enhances the robustness of the object removal process.
    If multiple candidates pass all filtering stages, the one with the lowest multimodal CLIP score is selected.
     Prior to choosing the CLIP Consensus and Multimodal CLIP filters thresholds, we manually annotated 500 inpainted images, classifying them as successful or failed removals. We tested the filters across varying thresholds and plotted the percentage of successful inpainted images against the percentage of filtered images. As shown in \cref{SM_figure:Concensus_treshold} and \cref{SM_figure:Multimodal_treshold}, as the filters become more aggressive (lower thresholds), the proportion of successful inpainted images increases for both strategies. This implies that both filtering approaches effectively achieve their aim of filtering out unsuccessful inpainting outputs.  We selected thresholds where the slope of successful inpainting begins to plateau, minimizing the loss of images while maximizing quality.

    \item \underline{Consistency Enforcement}:   
    We aim to produce image targets that are consistent with the source ones.
    By conducting $\alpha$-blending between the source and inpainted image using the object mask, we limit differences to the mask area while ensuring a smooth, natural transition between regions (see example in \cref{SM_figure:consistency_enforcment}).
    \item \underline{Importance Filtering}: 
    In the final removal pipeline step, we filter out instances where the removed object has marginal semantic importance, as such edits are unlikely to be user-requested. 
    We use a CLIP image encoder to assess the similarity between source and target images—not limited to the object region—filtering cases exceeding a manually set threshold.
\end{itemize}

\vspace{-0.18cm}
\subsection{Generating Object Addition Instructions}
\label{section:instruction_generation}
\vspace{-0.1cm}
The \pipe{} dataset is designed to include triplets of source and target images, along with corresponding editing instructions in natural language.
However, the process outlined in \Cref{sec:method:dataset} only produces pairs of images and the raw class name of the object of interest.
To address this gap, we introduce three different strategies for enhancing our dataset with instructions:

\noindent
\textbf{Class name-based instructions. }
We augment raw object classes into object addition instructions using the format ``\texttt{add a <class>}'', leading to simple and concise instructions.

\noindent
\textbf{VLM-LLM based instructions. }
We propose an automatic procedure designed to produce more varied and comprehensive instructions than those based on class names.
Leveraging recent VLM and LLM advances, we craft instructions using a two-stage process, as illustrated in \Cref{method}.
In the first stage, we mask out non-object regions and insert the devised image into a VLM, namely CogVLM\footnote{
\url{https://hf.co/THUDM/cogvlm-chat-hf}
}~\citep{wang2024cogvlm}, prompting it to generate a detailed object caption that includes visual object details and fine-grained attributes.
In the second stage, the caption is reformatted into an instruction using the in-context learning (ICL) capabilities of the LLM.
Specifically, we utilize Mistral-7B~\citep{jiang2023mistral} with $5$ ICL examples of the required outputs, prompting it to generate instructions of varying lengths and complexity.
This two-stage process, designed to mitigate hallucinations frequently encountered with VLMs \citep{liu2024survey}, has been empirically validated as effective and is inspired by research demonstrating that breaking down tasks into specific model roles enhances LLMs performance \citep{wang2024survey}.
Further details of this procedure are provided in the supplementary materials.

\noindent
\textbf{Manual Reference-based Instructions. }
To enrich our dataset with additional nuanced, compositional object details, we utilize three object reference datasets: RefCOCO, RefCOCO+~\citep{kazemzadeh2014referitgame}, and RefCOCOg~\citep{mao2016generation}.
We transform the references into instructions using the template: “\texttt{add a <object reference>}”, where “\texttt{<object reference>}” is replaced with the dataset's object description.

\begin{figure}[t]
  \centering
  \includegraphics[width=0.9\linewidth]{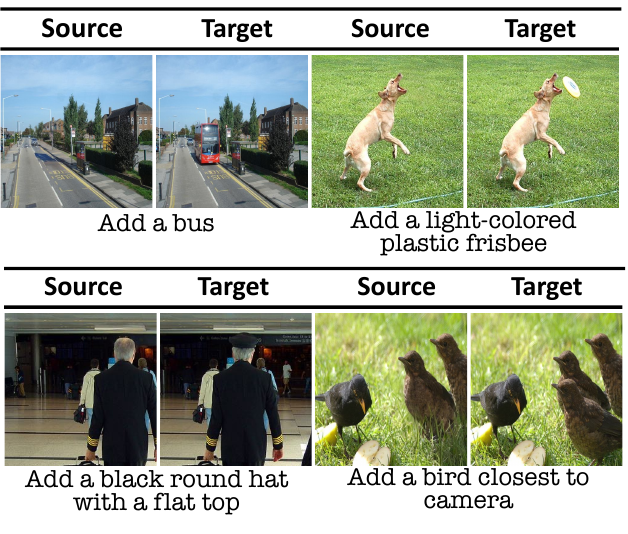}
   \vspace{-0.45cm}
  \caption{\textbf{
  \pipe{} dataset Examples.}
    Samples from \pipe{} using different instruction generation techniques: class name-based (left), VLM-LLM based (center), and reference-based (right).
  }
  \label{Fig:Dataset_Examples}
 \vspace{-0.4cm}
\end{figure}

Incorporating these diverse approaches produces $1,879,919$ different realistic object addition instructions, encompassing both concise and detailed editing scenarios.
Examples from \pipe{} using these diverse approaches are presented in \Cref{Fig:Dataset_Examples} and the appendix.
In \Cref{Table:datasets}, \pipe{} is compared with other image editing datasets.
It sets a new benchmark in image and editing instruction count by a significant margin.
Notably, it is the only dataset offering real target images and class diversity.

\vspace{-0.1cm}
\section{Model Training}\label{section:model_train}
\vspace{-0.15cm}
We detail the methodology used to train an image editing model using the proposed dataset, as illustrated in orange in \Cref{method}.
We leverage the SD 1.5 model~\citep{rombach2022high} for both its architecture and initial weights. 
This text-conditioned diffusion model incorporates a pre-trained variational autoencoder and a U-Net~\citep{ronneberger2015unet}, which is responsible for the diffusion denoising within the latent space of the former.
We denote the model parameters as $\theta$, the noisy latent variable at timestep $t$ as $z_t$, and the corresponding score estimate as $e_\theta$.
Similar to SD, our editing process is conditioned on a textual instruction encoding $c_T$ through cross-attention which integrates text encodings with visual representations.
We employ classifier-free guidance (CFG)~\citep{ho2022classifier} to enhance alignment between the output image and the instruction encoding $c_T$.
Contrary to SD, which generates a completely new image, our method involves editing an existing one.
Thus, similarly to IP2P, we condition the diffusion process not only on $c_T$ but also on the input image, denoted as $c_I$.
Liu et al.~\citep{liu2022compositional} demonstrated that a diffusion model can be conditioned on multiple targets, adapting CFG accordingly.
Using CFG necessitates modeling both conditional and unconditional scores.
To facilitate this, during training we set $c_T=\varnothing$ with probability $p=0.05$  (no text conditioning), $c_I = \varnothing$ with $p=0.05$ (no image conditioning), and $c_I = \varnothing,c_T=\varnothing $ with $p=0.05$ (no conditioning).
During inference, using CFG, we compute the score considering both the instruction and the source image. Further implementation details and hyperparameters are provided in the appendix.

\begin{table}[t]
\setlength\tabcolsep{0pt}
\vspace{-0.15cm}
\begin{tabular*}{1.03\linewidth}{@{\extracolsep{\fill}}lcccccc} 
\toprule

{\textbf{Dataset}} & { Real} & {Real} & {General} & {$\#$} & {$\#$}\\
\vspace{-13pt} \\
{} & {Source} & {Target} & {Classes}  & {Images} & {Edits}\\
{} & {Images} & {Images} & {}  & { } & { }\\

\midrule
Oxford-Flower  & \ding{51} & \ding{51} & \ding{55} & 8,189 & 8,189\\ 
CUB-Bird  & \ding{51} & \ding{51} & \ding{55} & 11,788 & 11,788\\
EditBench  &
{\ding{51}}\raisebox{-0.9ex}{\textsuperscript{\kern-0.9em\scalebox{1.6}{\ding{55}}}}
& \textbf{--} & \ding{51} & 240 & 960\\
InstructPix2Pix & \ding{55} & \ding{55} & \ding{51} & 313,010 & 313,010 \\ 
MagicBrush  & \ding{51} & 
{\ding{51}}\raisebox{-0.9ex}{\textsuperscript{\kern-0.9em\scalebox{1.6}{\ding{55}}}}
& \ding{51}  & 10,388 & 10,388\\ 
\midrule
\pipe{}  & 
{\ding{51}}\raisebox{-0.9ex}{\textsuperscript{\kern-0.9em\scalebox{1.6}{\ding{55}}}}
& \ding{51} & \ding{51} & 889,230 & 1,879,919\\
\bottomrule 
\end{tabular*}
\vspace{-0.15cm}
\caption{
\textbf{Datasets Comparison.}
Review of \pipe{} with other editing datasets.
\ding{51} signifies fulfillment,
\ding{55} indicates non-fulfillment,
and {\ding{51}\raisebox{-0.9ex}{\textsuperscript{\kern-0.9em\scalebox{1.6}{\ding{55}}}}} denotes partial fulfillment,
where images are real outside inpainted areas.
"\textbf{--}" means no such images available.
"General Classes" indicates dataset class diversity.
}
\label{Table:datasets}
\vspace{-0.4cm}
\end{table}

\section{Experiments}\label{sec:exps}
Image editing can yield countless different valid outcomes, making its evaluation a significant challenge. To address this, we perform a diverse array of experiments. Given that \pipe{} is primarily designed for object addition, we initially focus our experiments on this task before extending its application to general editing (in \Cref{section:ip2p_merge}). We quantitatively and qualitatively compare our model with top-performing methods, complemented by an in-depth detailed human evaluation survey. Additionally, in the appendix, we include an ablation study of the VLM-LLM pipeline.

\vspace{-0.1cm}
\subsection{Experimental Settings}
\vspace{-0.15cm}
\label{sec:setting}
We consider three benchmarks to evaluate our model's capabilities in object addition --
(i) \pipe{} test set: 750 images from the COCO validation split, generated using the pipeline outlined in \Cref{sec:dataset}.
(ii) OPA~\citep{liu2021opa}: An object placement assessment dataset that includes source and target images, along with objects to be added.
(iii) MagicBrush~\citep{zhang2024magicbrush}: A partially synthetic image editing benchmark comprising training and testing sets.
To evaluate object addition, we automatically filter the dataset for this task (details in the appendix), resulting in a $144$ edits subset.

\vspace{-0.1cm}
\subsection{Quantitative Evaluation}
\vspace{-0.15cm}
\label{sec:res1}

We compare our model with leading image editing models, including Hive~\citep{zhang2023hive}, IP2P~\citep{brooks2023instructpix2pix}, VQGAN-CLIP~\citep{crowson2022vqganclip}, SDEdit~\citep{meng2021sdedit}, Null-Text-Inversion~\citep{mokady2023null}, Pix2PixZero~\citep{parmar2023zero} and Edit-Freindly DDPM~\citep{huberman2024edit}.
For evaluating objects additions, we use the standardized metrics from MagicBrush~\citep{zhang2024magicbrush}.
These metrics compare edited outcomes to ground-truth targets using both
model-free ($L_1$ and $L_2$ distances) and model-based (CLIP~\citep{radford2021learning} and DINO~\citep{caron2021emerging} embedding cosine distances) measures.
Model-free metrics penalize global changes affecting non-object regions, while model-based approaches evaluate overall semantic similarity. 
When the edited target caption is available, we use CLIP-T~\citep{ruiz2023dreambooth} to measure its alignment with the edited image.
To complement our evaluation, we adopt the recently proposed Conditional Maximum Mean Discrepancy (CMMD) metric~\citep{jayasumana2024rethinking}. Like the popular Fréchet Inception Distance (FID)~\citep{heusel2017gans}, this metric measures the distributional distance between groups of images. However, unlike FID, CMMD uses CLIP embeddings and works effectively with a reduced number of samples, enabling us to measure distribution distances for small datasets like MagicBrush.
To further demonstrate the superiority of our model, we adopt a measure utilized by~\citep{brooks2023instructpix2pix}.
This measure, using changing image guidance scales ($s_I$), plots a  graph of two metrics of the edited outcome, both independent of a ground-truth target image:
(i) CLIP similarity with the input image.
(ii) Directional CLIP similarity~\citep{gal2022stylegan}, which evaluates changes between source-target image embeddings and source-target text caption embeddings.
This plot presents a trade-off between preserving the original content and achieving the desired edits.

\begin{table}[t]
        \vspace{-0.3cm}
        \hspace{-0.15cm}
        \begin{tabular*}{1.04\linewidth}{@{\extracolsep{\fill}}l@{\hspace{0.6em}}c@{\hspace{0.6em}}c@{\hspace{0.2em}}c@{\hspace{0.2em}}c@{\hspace{0.2em}}c@{\hspace{0.2em}}c}
            \toprule 
            \textbf{Methods} & L1$_\downarrow$ & L2$_\downarrow$ & CLIP-I$_\uparrow$ & DINO$_\uparrow$ & CLIP-T$_\uparrow$ & CMMD$_\downarrow$ \\
            \midrule
            VQGAN-CLIP  & .211 & .078 & .670 & .507 & \textbf{.484} & .862 \\
            SDEdit  & .168 & .057 & .765 & .572 & .325 & .539 \\
            Null-Text-Inv & \textbf{.072} & \textbf{.017} & .877 & .817 & .299 & .303 \\
            Pix2PixZero  & .086 & .024 & .846 & .750 & .294 & .322 \\
            EF-DDPM  & .110 & .030 & .844 & .716 & .328 &  .342\\
            Hive  & .095 & .026 & .846 & .782 & .297 & .353 \\
            IP2P  & .100 & .031 & .860 & .766 & .289 & .363 \\
            Ours & \textbf{.072} & .025 & \textbf{.900} & \textbf{.852} & .302 & \textbf{.301} \\
            \midrule
            \multicolumn{7}{c}{Fine-tune on MagicBrush}\\
            \midrule
            IP2P  & .077 & .028 & .902 & .867 & .306 &  .352\\
            Ours  & \textbf{.067} & \textbf{.023} & \textbf{.910} & \textbf{.897} & \textbf{.308} &  \textbf{.298}\\
            \bottomrule 
        \end{tabular*}
        \vspace{-0.2cm}
        \caption{\textbf{Results on MagicBrush} \underline{Top}: Our model and various baselines tested on the MagicBrush test set subset. \underline{Bottom}: Our model and IP2P fine-tuned on MagicBrush.}
        \label{Table:MagicBrush}
\end{table}

\begin{table}[t!]
    \hfill
    \begin{minipage}{1\linewidth}
        \vspace{+0.25cm}
        \begin{tabular*}{1\linewidth}{@{\extracolsep{\fill}}l@{\hspace{0.6em}}c@{\hspace{0.6em}}c@{\hspace{0.3em}}c@{\hspace{0.3em}}c@{\hspace{0.3em}}c} 
            \toprule
            {\textbf{Methods}} & {L1$_\downarrow$} & {L2$_\downarrow$} & {CLIP-I$_\uparrow$} & {DINO$_\uparrow$} & {CMMD$_\uparrow$}  \\
            \midrule
            Hive & .088 & .021 &  .849 & .754 & .232 \\
            IP2P & .098 & .027  & .861 & .753 & .142\\ 
            Ours  & \textbf{.057} & \textbf{.014} & \textbf{.945} & \textbf{.903}& \textbf{.060} \\
            \bottomrule 
        \end{tabular*}
        \vspace{-0.3cm}
        \caption{\textbf{Results on \pipe{} Test Set.}}
        \vspace{0.2cm}
        \label{Table:Ours}
        \begin{tabular*}{1\linewidth}{@{\extracolsep{\fill}}l@{\hspace{0.6em}}c@{\hspace{0.6em}}c@{\hspace{0.3em}}c@{\hspace{0.3em}}c@{\hspace{0.3em}}c} 
            \toprule
            {\textbf{Methods}} & {L1$_\downarrow$} & {L2$_\downarrow$} & {CLIP-I$_\uparrow$} & {DINO$_\uparrow$} & {CMMD$_\uparrow$}\\
            \toprule
            Hive & .126 & .041 & .802 & .670 & .481  \\
            IP2P & .109 & .035  & .806 & .647 & .467 \\ 
            Ours  & \textbf{.084} & \textbf{.027} & \textbf{.848} & \textbf{.735}& \textbf{.360} \\
            \bottomrule 
        \end{tabular*}
        \vspace{-0.3cm}
        \caption{\textbf{Results on OPA.}}
        \label{Table:OPA}
    \end{minipage}
    \vspace{-0.5cm}
\end{table}

\vspace{-0.15cm}
\paragraph{\pipe{} Test Results.}
We evaluate our model against instruction-following models, Hive and IP2P, using the \pipe{} held-out test set and report the results in \Cref{Table:Ours}.
Our model significantly surpasses the baselines in $L_1$ and $L_2$ metrics, confirming its high consistency, and exhibits a higher level of semantic resemblance to the target ground truth image, as reflected in the CLIP-I and DINO scores.

\vspace{-0.15cm}
\paragraph{OPA Results.}
In \Cref{Table:OPA}, we evaluate our model on the OPA dataset. Our approach achieves the highest performance across all evaluated metrics.

\begin{table}[!ht]
        \vspace{-0.35cm}
        \includegraphics[width=\linewidth]{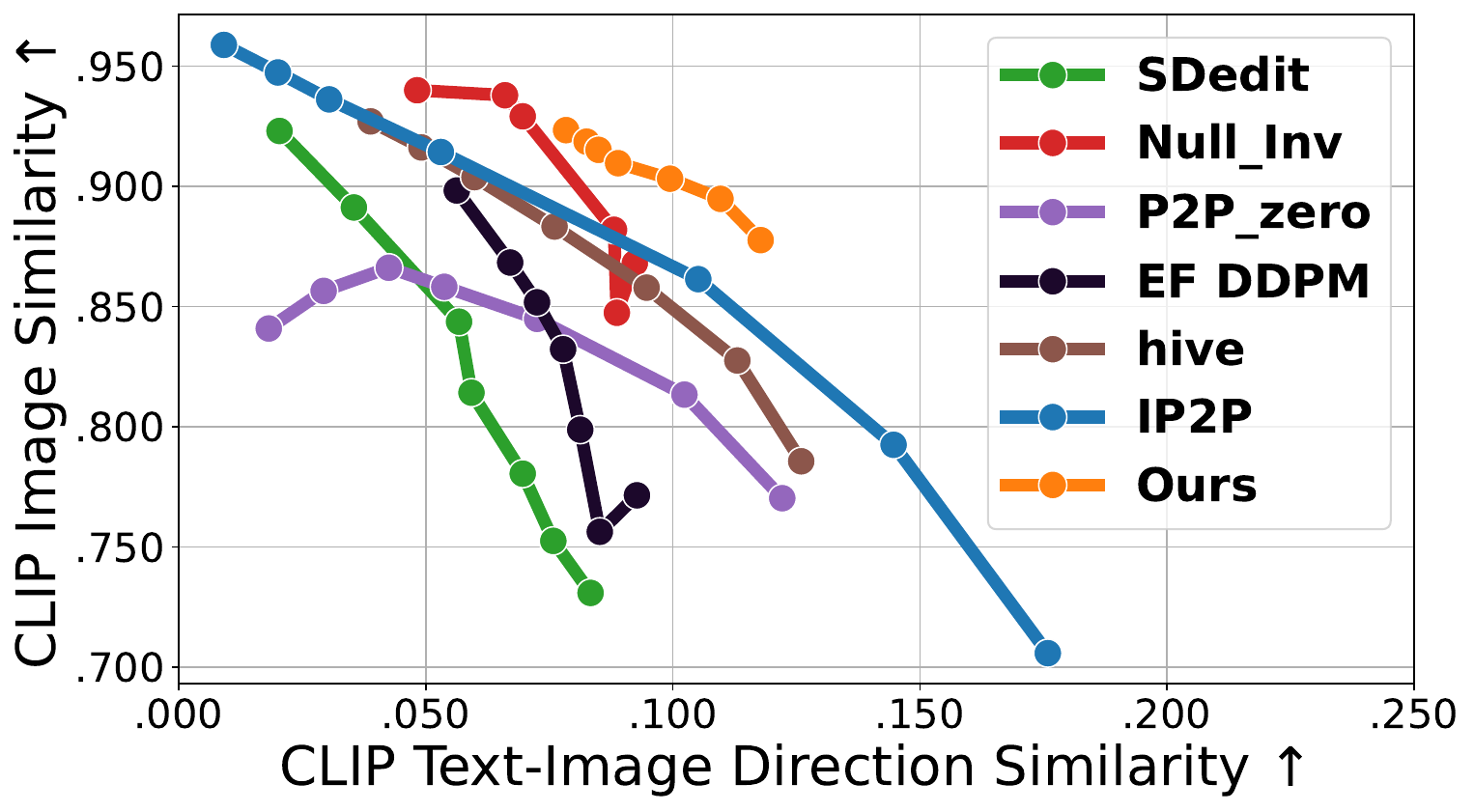}
        \vspace{-0.7cm}
        \captionof{figure}{\textbf{Consistency-Instruction Trade-off on MagicBrush.}}
        \vspace{-0.25cm}

        \label{fig:comp_real}
\end{table}

\paragraph{MagicBrush Results.}\label{pargagrh:mb}
We evaluate our model on the MagicBrush test subset, which includes source and target prompts in addition to instructions.
This allows us to compare our performance not only with instruction-following models like Hive and IP2P but also with prompt-based models like VQGAN-CLIP and SDEdit.
As presented in \Cref{Table:MagicBrush}, our model achieves the best results in most target image similarity metrics ($L_1$, CLIP-I, DINO and CMMD).

The target prompts also allow us to compare the CLIP-T metric.
While our model surpasses most methods in this metric, VQGAN-CLIP significantly outperforms it.
This result is expected as the latter maximizes an equivalent objective during the editing process. Although some methods outperform ours in CLIP-T, they fall behind in other metrics. To highlight our model's superior balance between consistency with the original image and following the instruction, we present comparisons in \cref{fig:comp_real}. As shown, our method outperforms all others in this tradeoff.
Following \citep{zhang2024magicbrush}, we also fine-tuned our model on the object-addition training subset of MagicBrush and compared it against the similarly fine-tuned IP2P, with our model exceeding IP2P in all metrics. Evaluations across the benchmarks show our model consistently outperforms competitors, affirming not only its high-quality outputs but also its robustness and adaptability across varied domains.

\vspace{-0.1cm}
\subsection{Qualitative Examples}
\vspace{-0.1cm}
Fig.~\ref{Fig:Visual_Comparison} qualitatively compares our model with other top-performing models across several datasets.
The results illustrate how the proposed model, in contrast to competing approaches, seamlessly adds synthesized objects into images naturally and coherently, while maintaining consistency with the original images before editing.
Furthermore, the examples, along with those in \Cref{Fig:Visual_Result_Teaser}, demonstrate our model's ability to generalize beyond its training classes, successfully integrating items such as a ''princess'' and ''buttoned shirt''. More examples are shown in the appendix.

\paragraph{Multiple Object Addition}
In \cref{sec:multiple_objects} of the appendix, we demonstrate how our method can be extended to add multiple objects.

\begin{table}[ht]

\vspace{-0.35cm}
\begin{tabular*}{1\linewidth}{
    @{\extracolsep{\fill}}l
    @{\hspace{0.4em}} c
    @{\hspace{0.6em}} c
    @{\hspace{0.2em}} c
    @{\hspace{0.2em}} c
    @{\hspace{0.2em}} c
    } 
    \toprule
    {\textbf{Methods}} & {L1$_\downarrow$} & {L2$_\downarrow$} & {CLIP-I$_\uparrow$} & {DINO$_\uparrow$} & {CLIP-T$_\uparrow$}
    \\
    \midrule
    IP2P  & .112 & .037  & .842 & .745 & .291 \\ 
    IP2P FT & .082 & .032  & .896 & .845 & .301\\
    Ours+IP2P FT & \textbf{.074} & \textbf{.026} & \textbf{.906} & \textbf{.866} & \textbf{.303}\\
    \bottomrule 
\end{tabular*}
\caption{\textbf{General Editing Results on MagicBrush Test Set.} Model performance evaluation on the full general editing MagicBrush test set.
The model, trained on the combined \pipe{} and IP2P dataset and fine-tuned on the MagicBrush training set, surpasses the previously top-performing fine-tuned IP2P,
demonstrating the potential of \pipe{} for enhancing general editing performance.}
\vspace{-0.15cm}
\label{Table:SP_Full_Brush}
\end{table}

\subsection{Qualitative Evaluation}\label{sec:res2}
\vspace{-0.1cm}
To complement the quantitative analysis, we conduct a human evaluation survey, comparing our model to IP2P.
To this end, we randomly sample $100$ images from the Conceptual Captions dataset~\citep{sharma2018conceptual} and request human annotators to provide reasonable addition instructions.
Next, we perform the edits using both models and request a different set of human evaluators to review their success.
We adopt the queries from \citep{zhang2024magicbrush} and ask evaluators to assess two aspects: alignment faithfulness between results and edit requests, and the output's general quality and consistency.
Overall, we collected $1,833$ individual responses from $57$ different human evaluators, all participants from a pool of random internet users.
To minimize biases and ensure an impartial evaluation, they completed the survey
unaware of the research goals.
We quantify edit faithfulness and output quality using two metrics: (i) overall global preference measured in percentage and (ii) aggregated per-image preference in absolute numbers (summed to $100$).
The results in \Cref{Table:human} showcase a substantial preference by human observers for our model’s outputs in both following instructions and image quality.
On average, the global preference metric indicates that our model is preferred approximately $72.6\%$  of the time.
Additional survey details are provided in the supplementary materials. An additional human evaluation against hive is presented in \cref{SM_table:human_hive}.

\begin{table}[t]
\setlength\tabcolsep{0pt}

\vspace{-0.1cm}
\begin{tabular*}{1\linewidth}{@{\extracolsep{\fill}}lcccc} 
\toprule
\multirow{2}{*}{\textbf{   Methods   }} & \multicolumn{2}{c}{\textbf{Edit faithfulness}} & \multicolumn{2}{c}{\textbf{Quality}} \\
& Overall [\%] & Per-image & Overall [\%] & Per-image \\
\midrule
IP2P & 26.4          & 28          & 28.5          & 31 \\ 
Ours                                             & \textbf{73.6} & \textbf{72} & \textbf{71.5} & \textbf{69} \\
\bottomrule 
\end{tabular*}
\caption{\textbf{Human Evaluation.} Comparison of our model with IP2P on edit faithfulness and quality. ``Overall'' represents the total vote percentage. ``Per-image'' quantifies the number of images where a model's outputs were preferred.}
\label{Table:human}
\vspace{-0.4cm}
\end{table}

\begin{table}
\vspace{-0.35cm}
\centering
\includegraphics[width=0.88\linewidth]{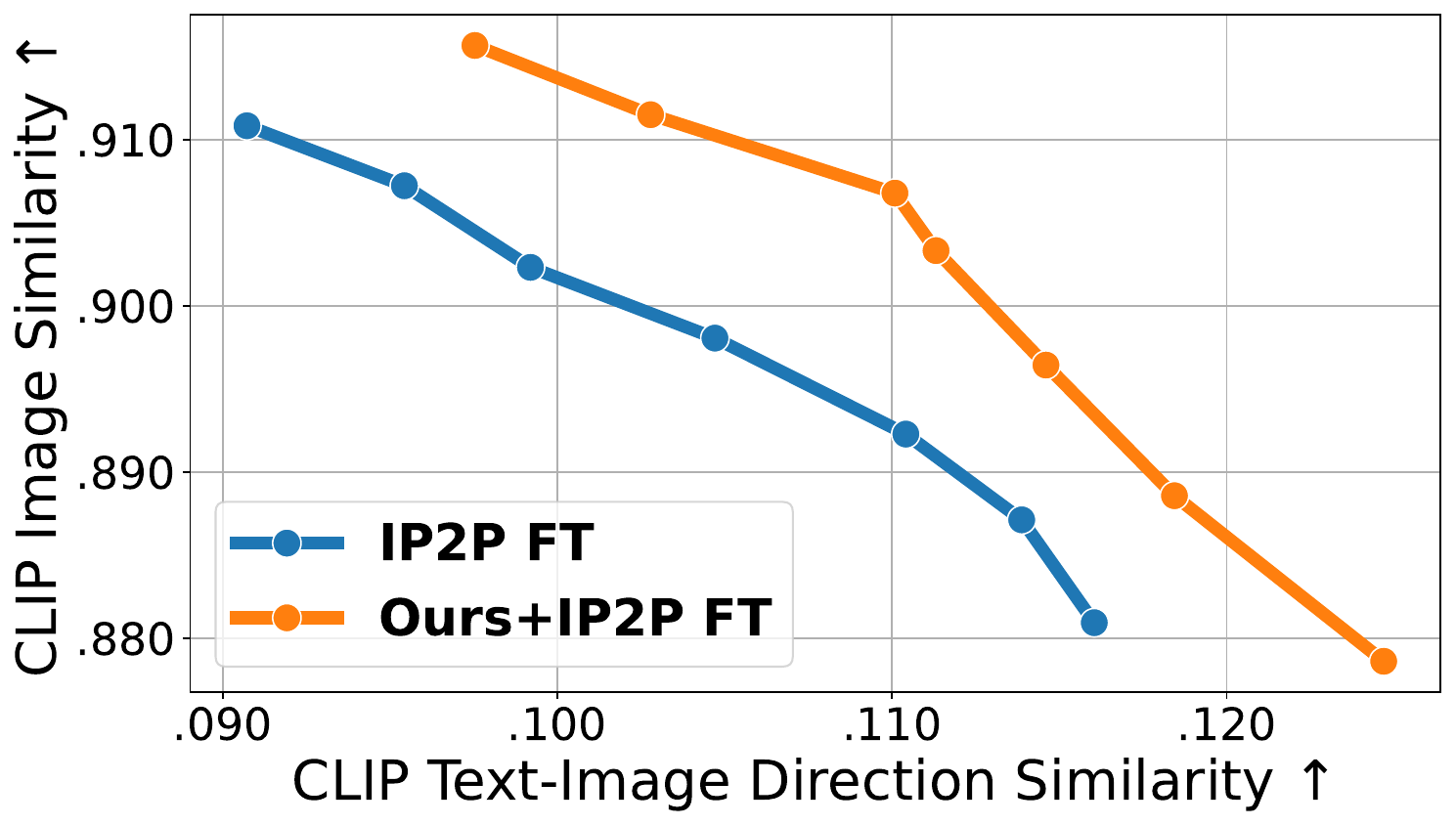}
\captionof{figure}{\textbf{General Editing Consistency-Instruction Trade-off.} Trade-off between consistency to input image (Y-axis) and edit adherence (X-axis),
with text guidance fixed at $7$ and varying image guidance in the range $[1, 2.5]$.}
\vspace{-0.2cm}
\label{Fig:general_editting_plot}
\end{table}

\section{Leveraging PIPE for General Editing}\label{section:ip2p_merge} 

We explore the application of our dataset in the broader context of image editing, extending its use beyond merely object addition.
We combine the IP2P general editing dataset with \pipe{} and use it to train an editing diffusion model, following the procedure outlined in \Cref{section:model_train}.
For evaluation, we utilized the entire MagicBrush test set, comparing our model against the IP2P model, both with and without MagicBrush fine-tuning.
Diverging from the object addition concentrated approach, 
the model is fine-tuned using the full MagicBrush training set.

To ensure fairness and reproducibility, all models were run with the same seed.
Evaluations were conducted using the script provided by \citep{zhang2024magicbrush}, and the official models were employed with their recommended inference parameters.
As illustrated in \Cref{Table:SP_Full_Brush}, our model sets new state-of-the-art scores for the general editing task, surpassing the current leading models.
As presented in~\Cref{Fig:general_editting_plot}, our fine-tuned model surpasses the current leading IP2P fine-tuned model, demonstrating higher image consistency for the same directional similarity values.
The results collectively affirm that the \pipe{} dataset can be combined with any editing dataset and improve overall performance.
In the appendix, we provide a qualitative visual comparison, showcasing the enhanced capabilities of the new model, not limited
to object addition, as well as similar plots for the object addition subset used in \Cref{sec:exps}.

\vspace{-0.23cm}
\section{Discussion}
\vspace{-0.16cm}
\label{sec:discussion}
In this work, we introduce the Paint by Inpaint framework, which identifies and leverages the fact that adding objects to images is fundamentally the inverse process of removing them.
Building on this insight, by harnessing the wealth of available segmentation datasets and utilizing a high-performance mask-based inpainting model, we present \pipe{}, an object addition dataset.
Unlike other mask-free, instruction-following editing datasets, \pipe{} is both large-scale and features consistent and natural editing target images.
We demonstrate that training a diffusion model on the dataset leads to state-of-the-art performance in instruction-based image editing, proving the value of the PIPE dataset in achieving consistent and realistic image edits.

{
    \small
    \bibliographystyle{ieeenat_fullname}
    \bibliography{main}

\begin{thebibliography}{75}
\providecommand{\natexlab}[1]{#1}
\providecommand{\url}[1]{\texttt{#1}}
\expandafter\ifx\csname urlstyle\endcsname\relax
  \providecommand{\doi}[1]{doi: #1}\else
  \providecommand{\doi}{doi: \begingroup \urlstyle{rm}\Url}\fi

\bibitem[Avrahami et~al.(2022)Avrahami, Lischinski, and Fried]{avrahami2022blended}
Omri Avrahami, Dani Lischinski, and Ohad Fried.
\newblock Blended diffusion for text-driven editing of natural images.
\newblock In \emph{Proceedings of the IEEE/CVF Conference on Computer Vision and Pattern Recognition}, pages 18208--18218, 2022.

\bibitem[Bai et~al.(2023)Bai, Bai, Yang, Wang, Tan, Wang, Lin, Zhou, and Zhou]{bai2023qwen}
Jinze Bai, Shuai Bai, Shusheng Yang, Shijie Wang, Sinan Tan, Peng Wang, Junyang Lin, Chang Zhou, and Jingren Zhou.
\newblock Qwen-vl: A frontier large vision-language model with versatile abilities.
\newblock \emph{arXiv preprint arXiv:2308.12966}, 2023.

\bibitem[Bar-Tal et~al.(2022)Bar-Tal, Ofri-Amar, Fridman, Kasten, and Dekel]{bar2022text2live}
Omer Bar-Tal, Dolev Ofri-Amar, Rafail Fridman, Yoni Kasten, and Tali Dekel.
\newblock Text2live: Text-driven layered image and video editing.
\newblock In \emph{European conference on computer vision}, pages 707--723. Springer, 2022.

\bibitem[Brooks et~al.(2023)Brooks, Holynski, and Efros]{brooks2023instructpix2pix}
Tim Brooks, Aleksander Holynski, and Alexei~A Efros.
\newblock Instructpix2pix: Learning to follow image editing instructions.
\newblock In \emph{Proceedings of the IEEE/CVF Conference on Computer Vision and Pattern Recognition}, pages 18392--18402, 2023.

\bibitem[Brown et~al.(2020)Brown, Mann, Ryder, Subbiah, Kaplan, Dhariwal, Neelakantan, Shyam, Sastry, Askell, et~al.]{brown2020language}
Tom Brown, Benjamin Mann, Nick Ryder, Melanie Subbiah, Jared~D Kaplan, Prafulla Dhariwal, Arvind Neelakantan, Pranav Shyam, Girish Sastry, Amanda Askell, et~al.
\newblock Language models are few-shot learners.
\newblock \emph{Advances in neural information processing systems}, 33:\penalty0 1877--1901, 2020.

\bibitem[Canberk et~al.(2024)Canberk, Bondarenko, Ozguroglu, Liu, and Vondrick]{canberk2024erasedraw}
Alper Canberk, Maksym Bondarenko, Ege Ozguroglu, Ruoshi Liu, and Carl Vondrick.
\newblock Erasedraw: Learning to insert objects by erasing them from images.
\newblock In \emph{European Conference on Computer Vision}, pages 144--160. Springer, 2024.

\bibitem[Caron et~al.(2021)Caron, Touvron, Misra, J{\'e}gou, Mairal, Bojanowski, and Joulin]{caron2021emerging}
Mathilde Caron, Hugo Touvron, Ishan Misra, Herv{\'e} J{\'e}gou, Julien Mairal, Piotr Bojanowski, and Armand Joulin.
\newblock Emerging properties in self-supervised vision transformers.
\newblock In \emph{Proceedings of the IEEE/CVF international conference on computer vision}, pages 9650--9660, 2021.

\bibitem[Chen et~al.(2024)Chen, Hu, Li, Ruiz, Jia, Chang, and Cohen]{chen2024subject}
Wenhu Chen, Hexiang Hu, Yandong Li, Nataniel Ruiz, Xuhui Jia, Ming-Wei Chang, and William~W Cohen.
\newblock Subject-driven text-to-image generation via apprenticeship learning.
\newblock \emph{Advances in Neural Information Processing Systems}, 36, 2024.

\bibitem[Crowson et~al.(2022)Crowson, Biderman, Kornis, Stander, Hallahan, Castricato, and Raff]{crowson2022vqganclip}
Katherine Crowson, Stella Biderman, Daniel Kornis, Dashiell Stander, Eric Hallahan, Louis Castricato, and Edward Raff.
\newblock Vqgan-clip: Open domain image generation and editing with natural language guidance, 2022.

\bibitem[Dai et~al.(2023)Dai, Li, Li, Tiong, Zhao, Wang, Li, Fung, and Hoi]{dai2023instructblip}
Wenliang Dai, Junnan Li, Dongxu Li, Anthony Meng~Huat Tiong, Junqi Zhao, Weisheng Wang, Boyang Li, Pascale Fung, and Steven Hoi.
\newblock Instructblip: Towards general-purpose vision-language models with instruction tuning, 2023.

\bibitem[El-Nouby et~al.(2019)El-Nouby, Sharma, Schulz, Hjelm, Asri, Kahou, Bengio, and Taylor]{el2019tell}
Alaaeldin El-Nouby, Shikhar Sharma, Hannes Schulz, Devon Hjelm, Layla~El Asri, Samira~Ebrahimi Kahou, Yoshua Bengio, and Graham~W Taylor.
\newblock Tell, draw, and repeat: Generating and modifying images based on continual linguistic instruction.
\newblock In \emph{Proceedings of the IEEE/CVF International Conference on Computer Vision}, pages 10304--10312, 2019.

\bibitem[Gal et~al.(2021)Gal, Patashnik, Maron, Chechik, and Cohen-Or]{gal2021stylegan}
Rinon Gal, Or Patashnik, Haggai Maron, Gal Chechik, and Daniel Cohen-Or.
\newblock Stylegan-nada: Clip-guided domain adaptation of image generators.
\newblock \emph{arXiv preprint arXiv:2108.00946}, 2021.

\bibitem[Gal et~al.(2022{\natexlab{a}})Gal, Alaluf, Atzmon, Patashnik, Bermano, Chechik, and Cohen-Or]{gal2022image}
Rinon Gal, Yuval Alaluf, Yuval Atzmon, Or Patashnik, Amit~H Bermano, Gal Chechik, and Daniel Cohen-Or.
\newblock An image is worth one word: Personalizing text-to-image generation using textual inversion.
\newblock \emph{arXiv preprint arXiv:2208.01618}, 2022{\natexlab{a}}.

\bibitem[Gal et~al.(2022{\natexlab{b}})Gal, Patashnik, Maron, Bermano, Chechik, and Cohen-Or]{gal2022stylegan}
Rinon Gal, Or Patashnik, Haggai Maron, Amit~H Bermano, Gal Chechik, and Daniel Cohen-Or.
\newblock Stylegan-nada: Clip-guided domain adaptation of image generators.
\newblock \emph{ACM Transactions on Graphics (TOG)}, 41\penalty0 (4):\penalty0 1--13, 2022{\natexlab{b}}.

\bibitem[Ganz et~al.(2023)Ganz, Nuriel, Aberdam, Kittenplon, Mazor, and Litman]{ganz2023models}
Roy Ganz, Oren Nuriel, Aviad Aberdam, Yair Kittenplon, Shai Mazor, and Ron Litman.
\newblock Towards models that can see and read, 2023.

\bibitem[Ganz et~al.(2024)Ganz, Kittenplon, Aberdam, Avraham, Nuriel, Mazor, and Litman]{ganz2024question}
Roy Ganz, Yair Kittenplon, Aviad Aberdam, Elad~Ben Avraham, Oren Nuriel, Shai Mazor, and Ron Litman.
\newblock Question aware vision transformer for multimodal reasoning, 2024.

\bibitem[Gupta et~al.(2019)Gupta, Dollar, and Girshick]{gupta2019lvis}
Agrim Gupta, Piotr Dollar, and Ross Girshick.
\newblock Lvis: A dataset for large vocabulary instance segmentation.
\newblock In \emph{Proceedings of the IEEE/CVF conference on computer vision and pattern recognition}, pages 5356--5364, 2019.

\bibitem[Hertz et~al.(2022)Hertz, Mokady, Tenenbaum, Aberman, Pritch, and Cohen-Or]{hertz2022prompt}
Amir Hertz, Ron Mokady, Jay Tenenbaum, Kfir Aberman, Yael Pritch, and Daniel Cohen-Or.
\newblock Prompt-to-prompt image editing with cross attention control.
\newblock \emph{arXiv preprint arXiv:2208.01626}, 2022.

\bibitem[Heusel et~al.(2017)Heusel, Ramsauer, Unterthiner, Nessler, and Hochreiter]{heusel2017gans}
Martin Heusel, Hubert Ramsauer, Thomas Unterthiner, Bernhard Nessler, and Sepp Hochreiter.
\newblock Gans trained by a two time-scale update rule converge to a local nash equilibrium.
\newblock \emph{Advances in neural information processing systems}, 30, 2017.

\bibitem[Ho and Salimans(2022)]{ho2022classifier}
Jonathan Ho and Tim Salimans.
\newblock Classifier-free diffusion guidance.
\newblock \emph{arXiv preprint arXiv:2207.12598}, 2022.

\bibitem[Ho et~al.(2020)Ho, Jain, and Abbeel]{ho2020denoising}
Jonathan Ho, Ajay Jain, and Pieter Abbeel.
\newblock Denoising diffusion probabilistic models.
\newblock \emph{Advances in neural information processing systems}, 33:\penalty0 6840--6851, 2020.

\bibitem[Huberman-Spiegelglas et~al.(2024)Huberman-Spiegelglas, Kulikov, and Michaeli]{huberman2024edit}
Inbar Huberman-Spiegelglas, Vladimir Kulikov, and Tomer Michaeli.
\newblock An edit friendly ddpm noise space: Inversion and manipulations.
\newblock In \emph{Proceedings of the IEEE/CVF Conference on Computer Vision and Pattern Recognition}, pages 12469--12478, 2024.

\bibitem[Jayasumana et~al.(2024)Jayasumana, Ramalingam, Veit, Glasner, Chakrabarti, and Kumar]{jayasumana2024rethinking}
Sadeep Jayasumana, Srikumar Ramalingam, Andreas Veit, Daniel Glasner, Ayan Chakrabarti, and Sanjiv Kumar.
\newblock Rethinking fid: Towards a better evaluation metric for image generation.
\newblock In \emph{Proceedings of the IEEE/CVF Conference on Computer Vision and Pattern Recognition}, pages 9307--9315, 2024.

\bibitem[Jiang et~al.(2023)Jiang, Sablayrolles, Mensch, Bamford, Chaplot, Casas, Bressand, Lengyel, Lample, Saulnier, et~al.]{jiang2023mistral}
Albert~Q Jiang, Alexandre Sablayrolles, Arthur Mensch, Chris Bamford, Devendra~Singh Chaplot, Diego de~las Casas, Florian Bressand, Gianna Lengyel, Guillaume Lample, Lucile Saulnier, et~al.
\newblock Mistral 7b.
\newblock \emph{arXiv preprint arXiv:2310.06825}, 2023.

\bibitem[Kawar et~al.(2023)Kawar, Zada, Lang, Tov, Chang, Dekel, Mosseri, and Irani]{kawar2023imagic}
Bahjat Kawar, Shiran Zada, Oran Lang, Omer Tov, Huiwen Chang, Tali Dekel, Inbar Mosseri, and Michal Irani.
\newblock Imagic: Text-based real image editing with diffusion models.
\newblock In \emph{Proceedings of the IEEE/CVF Conference on Computer Vision and Pattern Recognition}, pages 6007--6017, 2023.

\bibitem[Kazemzadeh et~al.(2014)Kazemzadeh, Ordonez, Matten, and Berg]{kazemzadeh2014referitgame}
Sahar Kazemzadeh, Vicente Ordonez, Mark Matten, and Tamara Berg.
\newblock Referitgame: Referring to objects in photographs of natural scenes.
\newblock In \emph{Proceedings of the 2014 conference on empirical methods in natural language processing (EMNLP)}, pages 787--798, 2014.

\bibitem[Kulikov et~al.(2024)Kulikov, Kleiner, Huberman-Spiegelglas, and Michaeli]{kulikov2024flowedit}
Vladimir Kulikov, Matan Kleiner, Inbar Huberman-Spiegelglas, and Tomer Michaeli.
\newblock Flowedit: Inversion-free text-based editing using pre-trained flow models.
\newblock \emph{arXiv preprint arXiv:2412.08629}, 2024.

\bibitem[Kuznetsova et~al.(2020{\natexlab{a}})Kuznetsova, Rom, Alldrin, Uijlings, Krasin, Pont-Tuset, Kamali, Popov, Malloci, Kolesnikov, Duerig, and Ferrari]{OpenImages}
Alina Kuznetsova, Hassan Rom, Neil Alldrin, Jasper Uijlings, Ivan Krasin, Jordi Pont-Tuset, Shahab Kamali, Stefan Popov, Matteo Malloci, Alexander Kolesnikov, Tom Duerig, and Vittorio Ferrari.
\newblock The open images dataset v4: Unified image classification, object detection, and visual relationship detection at scale.
\newblock \emph{IJCV}, 2020{\natexlab{a}}.

\bibitem[Kuznetsova et~al.(2020{\natexlab{b}})Kuznetsova, Rom, Alldrin, Uijlings, Krasin, Pont-Tuset, Kamali, Popov, Malloci, Kolesnikov, et~al.]{kuznetsova2020open}
Alina Kuznetsova, Hassan Rom, Neil Alldrin, Jasper Uijlings, Ivan Krasin, Jordi Pont-Tuset, Shahab Kamali, Stefan Popov, Matteo Malloci, Alexander Kolesnikov, et~al.
\newblock The open images dataset v4: Unified image classification, object detection, and visual relationship detection at scale.
\newblock \emph{International Journal of Computer Vision}, 128\penalty0 (7):\penalty0 1956--1981, 2020{\natexlab{b}}.

\bibitem[Labs(2024)]{Flux2024}
Black~Forest Labs.
\newblock Flux.
\newblock \url{https://github.com/black-forest-labs/flux}, 2024.

\bibitem[Li et~al.(2023{\natexlab{a}})Li, Li, Savarese, and Hoi]{li2023blip}
Junnan Li, Dongxu Li, Silvio Savarese, and Steven Hoi.
\newblock Blip-2: Bootstrapping language-image pre-training with frozen image encoders and large language models.
\newblock \emph{arXiv preprint arXiv:2301.12597}, 2023{\natexlab{a}}.

\bibitem[Li et~al.(2023{\natexlab{b}})Li, Liu, Wu, Mu, Yang, Gao, Li, and Lee]{li2023gligen}
Yuheng Li, Haotian Liu, Qingyang Wu, Fangzhou Mu, Jianwei Yang, Jianfeng Gao, Chunyuan Li, and Yong~Jae Lee.
\newblock Gligen: Open-set grounded text-to-image generation.
\newblock In \emph{Proceedings of the IEEE/CVF Conference on Computer Vision and Pattern Recognition}, pages 22511--22521, 2023{\natexlab{b}}.

\bibitem[Lin et~al.(2014)Lin, Maire, Belongie, Hays, Perona, Ramanan, Doll{\'a}r, and Zitnick]{lin2014microsoft}
Tsung-Yi Lin, Michael Maire, Serge Belongie, James Hays, Pietro Perona, Deva Ramanan, Piotr Doll{\'a}r, and C~Lawrence Zitnick.
\newblock Microsoft coco: Common objects in context.
\newblock In \emph{Computer Vision--ECCV 2014: 13th European Conference, Zurich, Switzerland, September 6-12, 2014, Proceedings, Part V 13}, pages 740--755. Springer, 2014.

\bibitem[Lin et~al.(2024)Lin, Pathak, Li, Li, Xia, Neubig, Zhang, and Ramanan]{lin2024evaluating}
Zhiqiu Lin, Deepak Pathak, Baiqi Li, Jiayao Li, Xide Xia, Graham Neubig, Pengchuan Zhang, and Deva Ramanan.
\newblock Evaluating text-to-visual generation with image-to-text generation.
\newblock In \emph{European Conference on Computer Vision}, pages 366--384. Springer, 2024.

\bibitem[Liu et~al.(2023)Liu, Li, Li, and Lee]{liu2023improved}
Haotian Liu, Chunyuan Li, Yuheng Li, and Yong~Jae Lee.
\newblock Improved baselines with visual instruction tuning.
\newblock \emph{arXiv preprint arXiv:2310.03744}, 2023.

\bibitem[Liu et~al.(2024)Liu, Xue, Chen, Chen, Zhao, Wang, Hou, Li, and Peng]{liu2024survey}
Hanchao Liu, Wenyuan Xue, Yifei Chen, Dapeng Chen, Xiutian Zhao, Ke Wang, Liping Hou, Rongjun Li, and Wei Peng.
\newblock A survey on hallucination in large vision-language models.
\newblock \emph{arXiv preprint arXiv:2402.00253}, 2024.

\bibitem[Liu et~al.(2021)Liu, Liu, Zhang, Li, Niu, Liu, and Zhang]{liu2021opa}
Liu Liu, Zhenchen Liu, Bo Zhang, Jiangtong Li, Li Niu, Qingyang Liu, and Liqing Zhang.
\newblock Opa: object placement assessment dataset.
\newblock \emph{arXiv preprint arXiv:2107.01889}, 2021.

\bibitem[Liu et~al.(2022)Liu, Li, Du, Torralba, and Tenenbaum]{liu2022compositional}
Nan Liu, Shuang Li, Yilun Du, Antonio Torralba, and Joshua~B Tenenbaum.
\newblock Compositional visual generation with composable diffusion models.
\newblock In \emph{European Conference on Computer Vision}, pages 423--439. Springer, 2022.

\bibitem[Mao et~al.(2016)Mao, Huang, Toshev, Camburu, Yuille, and Murphy]{mao2016generation}
Junhua Mao, Jonathan Huang, Alexander Toshev, Oana Camburu, Alan~L Yuille, and Kevin Murphy.
\newblock Generation and comprehension of unambiguous object descriptions.
\newblock In \emph{Proceedings of the IEEE conference on computer vision and pattern recognition}, pages 11--20, 2016.

\bibitem[Meng et~al.(2021)Meng, He, Song, Song, Wu, Zhu, and Ermon]{meng2021sdedit}
Chenlin Meng, Yutong He, Yang Song, Jiaming Song, Jiajun Wu, Jun-Yan Zhu, and Stefano Ermon.
\newblock Sdedit: Guided image synthesis and editing with stochastic differential equations.
\newblock \emph{arXiv preprint arXiv:2108.01073}, 2021.

\bibitem[Mokady et~al.(2023)Mokady, Hertz, Aberman, Pritch, and Cohen-Or]{mokady2023null}
Ron Mokady, Amir Hertz, Kfir Aberman, Yael Pritch, and Daniel Cohen-Or.
\newblock Null-text inversion for editing real images using guided diffusion models.
\newblock In \emph{Proceedings of the IEEE/CVF Conference on Computer Vision and Pattern Recognition}, pages 6038--6047, 2023.

\bibitem[Nichol et~al.(2021)Nichol, Dhariwal, Ramesh, Shyam, Mishkin, McGrew, Sutskever, and Chen]{nichol2021glide}
Alex Nichol, Prafulla Dhariwal, Aditya Ramesh, Pranav Shyam, Pamela Mishkin, Bob McGrew, Ilya Sutskever, and Mark Chen.
\newblock Glide: Towards photorealistic image generation and editing with text-guided diffusion models.
\newblock \emph{arXiv preprint arXiv:2112.10741}, 2021.

\bibitem[Nilsback and Zisserman(2008)]{nilsback2008automated}
Maria-Elena Nilsback and Andrew Zisserman.
\newblock Automated flower classification over a large number of classes.
\newblock In \emph{2008 Sixth Indian conference on computer vision, graphics \& image processing}, pages 722--729. IEEE, 2008.

\bibitem[Oh et~al.(2001)Oh, Chen, Dorsey, and Durand]{oh2001image}
Byong~Mok Oh, Max Chen, Julie Dorsey, and Fr{\'e}do Durand.
\newblock Image-based modeling and photo editing.
\newblock In \emph{Proceedings of the 28th annual conference on Computer graphics and interactive techniques}, pages 433--442, 2001.

\bibitem[Ouyang et~al.(2022)Ouyang, Wu, Jiang, Almeida, Wainwright, Mishkin, Zhang, Agarwal, Slama, Ray, et~al.]{ouyang2022training}
Long Ouyang, Jeffrey Wu, Xu Jiang, Diogo Almeida, Carroll Wainwright, Pamela Mishkin, Chong Zhang, Sandhini Agarwal, Katarina Slama, Alex Ray, et~al.
\newblock Training language models to follow instructions with human feedback.
\newblock \emph{Advances in Neural Information Processing Systems}, 35:\penalty0 27730--27744, 2022.

\bibitem[Parmar et~al.(2023)Parmar, Kumar~Singh, Zhang, Li, Lu, and Zhu]{parmar2023zero}
Gaurav Parmar, Krishna Kumar~Singh, Richard Zhang, Yijun Li, Jingwan Lu, and Jun-Yan Zhu.
\newblock Zero-shot image-to-image translation.
\newblock In \emph{ACM SIGGRAPH 2023 Conference Proceedings}, pages 1--11, 2023.

\bibitem[Patashnik et~al.(2023)Patashnik, Garibi, Azuri, Averbuch-Elor, and Cohen-Or]{patashnik2023localizing}
Or Patashnik, Daniel Garibi, Idan Azuri, Hadar Averbuch-Elor, and Daniel Cohen-Or.
\newblock Localizing object-level shape variations with text-to-image diffusion models.
\newblock \emph{arXiv preprint arXiv:2303.11306}, 2023.

\bibitem[P{\'e}rez et~al.(2023)P{\'e}rez, Gangnet, and Blake]{perez2023poisson}
Patrick P{\'e}rez, Michel Gangnet, and Andrew Blake.
\newblock Poisson image editing.
\newblock In \emph{Seminal Graphics Papers: Pushing the Boundaries, Volume 2}, pages 577--582. 2023.

\bibitem[Pobitzer et~al.(2024)Pobitzer, Janicki, Rigotti, and Malossi]{pobitzer2024outline}
Markus Pobitzer, Filip Janicki, Mattia Rigotti, and Cristiano Malossi.
\newblock Outline-guided object inpainting with diffusion models.
\newblock \emph{arXiv preprint arXiv:2402.16421}, 2024.

\bibitem[Radford et~al.(2021)Radford, Kim, Hallacy, Ramesh, Goh, Agarwal, Sastry, Askell, Mishkin, Clark, et~al.]{radford2021learning}
Alec Radford, Jong~Wook Kim, Chris Hallacy, Aditya Ramesh, Gabriel Goh, Sandhini Agarwal, Girish Sastry, Amanda Askell, Pamela Mishkin, Jack Clark, et~al.
\newblock Learning transferable visual models from natural language supervision.
\newblock In \emph{International conference on machine learning}, pages 8748--8763. PMLR, 2021.

\bibitem[Ramesh et~al.(2022)Ramesh, Dhariwal, Nichol, Chu, and Chen]{ramesh2022hierarchical}
Aditya Ramesh, Prafulla Dhariwal, Alex Nichol, Casey Chu, and Mark Chen.
\newblock Hierarchical text-conditional image generation with clip latents.
\newblock \emph{arXiv preprint arXiv:2204.06125}, 1\penalty0 (2):\penalty0 3, 2022.

\bibitem[Rombach et~al.(2022)Rombach, Blattmann, Lorenz, Esser, and Ommer]{rombach2022high}
Robin Rombach, Andreas Blattmann, Dominik Lorenz, Patrick Esser, and Bj{\"o}rn Ommer.
\newblock High-resolution image synthesis with latent diffusion models.
\newblock In \emph{Proceedings of the IEEE/CVF conference on computer vision and pattern recognition}, pages 10684--10695, 2022.

\bibitem[Ronneberger et~al.(2015)Ronneberger, Fischer, and Brox]{ronneberger2015unet}
Olaf Ronneberger, Philipp Fischer, and Thomas Brox.
\newblock U-net: Convolutional networks for biomedical image segmentation, 2015.

\bibitem[Rotstein et~al.(2023)Rotstein, Bensaid, Brody, Ganz, and Kimmel]{rotstein2023fusecap}
Noam Rotstein, David Bensaid, Shaked Brody, Roy Ganz, and Ron Kimmel.
\newblock Fusecap: Leveraging large language models to fuse visual data into enriched image captions.
\newblock \emph{arXiv preprint arXiv:2305.17718}, 2023.

\bibitem[Rotstein et~al.(2024)Rotstein, Yona, Silver, Velich, Bensa{\"\i}d, and Kimmel]{rotstein2024pathways}
Noam Rotstein, Gal Yona, Daniel Silver, Roy Velich, David Bensa{\"\i}d, and Ron Kimmel.
\newblock Pathways on the image manifold: Image editing via video generation.
\newblock \emph{arXiv preprint arXiv:2411.16819}, 2024.

\bibitem[Ruiz et~al.(2023)Ruiz, Li, Jampani, Pritch, Rubinstein, and Aberman]{ruiz2023dreambooth}
Nataniel Ruiz, Yuanzhen Li, Varun Jampani, Yael Pritch, Michael Rubinstein, and Kfir Aberman.
\newblock Dreambooth: Fine tuning text-to-image diffusion models for subject-driven generation.
\newblock In \emph{Proceedings of the IEEE/CVF Conference on Computer Vision and Pattern Recognition}, pages 22500--22510, 2023.

\bibitem[Saharia et~al.(2022{\natexlab{a}})Saharia, Chan, Chang, Lee, Ho, Salimans, Fleet, and Norouzi]{saharia2022palette}
Chitwan Saharia, William Chan, Huiwen Chang, Chris Lee, Jonathan Ho, Tim Salimans, David Fleet, and Mohammad Norouzi.
\newblock Palette: Image-to-image diffusion models.
\newblock In \emph{ACM SIGGRAPH 2022 Conference Proceedings}, pages 1--10, 2022{\natexlab{a}}.

\bibitem[Saharia et~al.(2022{\natexlab{b}})Saharia, Chan, Saxena, Li, Whang, Denton, Ghasemipour, Gontijo~Lopes, Karagol~Ayan, Salimans, et~al.]{saharia2022photorealistic}
Chitwan Saharia, William Chan, Saurabh Saxena, Lala Li, Jay Whang, Emily~L Denton, Kamyar Ghasemipour, Raphael Gontijo~Lopes, Burcu Karagol~Ayan, Tim Salimans, et~al.
\newblock Photorealistic text-to-image diffusion models with deep language understanding.
\newblock \emph{Advances in Neural Information Processing Systems}, 35:\penalty0 36479--36494, 2022{\natexlab{b}}.

\bibitem[Sharma et~al.(2018)Sharma, Ding, Goodman, and Soricut]{sharma2018conceptual}
Piyush Sharma, Nan Ding, Sebastian Goodman, and Radu Soricut.
\newblock Conceptual captions: A cleaned, hypernymed, image alt-text dataset for automatic image captioning.
\newblock In \emph{Proceedings of the 56th Annual Meeting of the Association for Computational Linguistics (Volume 1: Long Papers)}, pages 2556--2565, 2018.

\bibitem[Song et~al.(2020)Song, Sohl-Dickstein, Kingma, Kumar, Ermon, and Poole]{song2020score}
Yang Song, Jascha Sohl-Dickstein, Diederik~P Kingma, Abhishek Kumar, Stefano Ermon, and Ben Poole.
\newblock Score-based generative modeling through stochastic differential equations.
\newblock \emph{arXiv preprint arXiv:2011.13456}, 2020.

\bibitem[Tarr{\'e}s et~al.(2024)Tarr{\'e}s, Lin, Zhang, Zhang, Song, Ruta, Gilbert, Collomosse, and Kim]{tarres2024thinking}
Gemma~Canet Tarr{\'e}s, Zhe Lin, Zhifei Zhang, Jianming Zhang, Yizhi Song, Dan Ruta, Andrew Gilbert, John Collomosse, and Soo~Ye Kim.
\newblock Thinking outside the bbox: Unconstrained generative object compositing.
\newblock \emph{arXiv preprint arXiv:2409.04559}, 2024.

\bibitem[Wah et~al.(2011)Wah, Branson, Welinder, Perona, and Belongie]{wah2011caltech}
Catherine Wah, Steve Branson, Peter Welinder, Pietro Perona, and Serge Belongie.
\newblock The caltech-ucsd birds-200-2011 dataset.
\newblock 2011.

\bibitem[Wang et~al.(2024{\natexlab{a}})Wang, Ma, Feng, Zhang, Yang, Zhang, Chen, Tang, Chen, Lin, et~al.]{wang2024survey}
Lei Wang, Chen Ma, Xueyang Feng, Zeyu Zhang, Hao Yang, Jingsen Zhang, Zhiyuan Chen, Jiakai Tang, Xu Chen, Yankai Lin, et~al.
\newblock A survey on large language model based autonomous agents.
\newblock \emph{Frontiers of Computer Science}, 18\penalty0 (6):\penalty0 1--26, 2024{\natexlab{a}}.

\bibitem[Wang et~al.(2023{\natexlab{a}})Wang, Saharia, Montgomery, Pont-Tuset, Noy, Pellegrini, Onoe, Laszlo, Fleet, Soricut, et~al.]{wang2023imagen}
Su Wang, Chitwan Saharia, Ceslee Montgomery, Jordi Pont-Tuset, Shai Noy, Stefano Pellegrini, Yasumasa Onoe, Sarah Laszlo, David~J Fleet, Radu Soricut, et~al.
\newblock Imagen editor and editbench: Advancing and evaluating text-guided image inpainting.
\newblock In \emph{Proceedings of the IEEE/CVF Conference on Computer Vision and Pattern Recognition}, pages 18359--18369, 2023{\natexlab{a}}.

\bibitem[Wang et~al.(2023{\natexlab{b}})Wang, Lv, Yu, Hong, Qi, Wang, Ji, Yang, Zhao, Song, et~al.]{wang2023cogvlm}
Weihan Wang, Qingsong Lv, Wenmeng Yu, Wenyi Hong, Ji Qi, Yan Wang, Junhui Ji, Zhuoyi Yang, Lei Zhao, Xixuan Song, et~al.
\newblock Cogvlm: Visual expert for pretrained language models.
\newblock \emph{arXiv preprint arXiv:2311.03079}, 2023{\natexlab{b}}.

\bibitem[Wang et~al.(2024{\natexlab{b}})Wang, Lv, Yu, Hong, Qi, Wang, Ji, Yang, Zhao, Song, Xu, Xu, Li, Dong, Ding, and Tang]{wang2024cogvlm}
Weihan Wang, Qingsong Lv, Wenmeng Yu, Wenyi Hong, Ji Qi, Yan Wang, Junhui Ji, Zhuoyi Yang, Lei Zhao, Xixuan Song, Jiazheng Xu, Bin Xu, Juanzi Li, Yuxiao Dong, Ming Ding, and Jie Tang.
\newblock Cogvlm: Visual expert for pretrained language models, 2024{\natexlab{b}}.

\bibitem[Xie et~al.(2023)Xie, Zhang, Lin, Hinz, and Zhang]{xie2023smartbrush}
Shaoan Xie, Zhifei Zhang, Zhe Lin, Tobias Hinz, and Kun Zhang.
\newblock Smartbrush: Text and shape guided object inpainting with diffusion model.
\newblock In \emph{Proceedings of the IEEE/CVF Conference on Computer Vision and Pattern Recognition}, pages 22428--22437, 2023.

\bibitem[Xu et~al.(2018)Xu, Zhang, Huang, Zhang, Gan, Huang, and He]{xu2018attngan}
Tao Xu, Pengchuan Zhang, Qiuyuan Huang, Han Zhang, Zhe Gan, Xiaolei Huang, and Xiaodong He.
\newblock Attngan: Fine-grained text to image generation with attentional generative adversarial networks.
\newblock In \emph{Proceedings of the IEEE conference on computer vision and pattern recognition}, pages 1316--1324, 2018.

\bibitem[Yildirim et~al.(2023)Yildirim, Baday, Erdem, Erdem, and Dundar]{yildirim2023inst}
Ahmet~Burak Yildirim, Vedat Baday, Erkut Erdem, Aykut Erdem, and Aysegul Dundar.
\newblock Inst-inpaint: Instructing to remove objects with diffusion models.
\newblock \emph{arXiv preprint arXiv:2304.03246}, 2023.

\bibitem[Yu et~al.(2023)Yu, Feng, Feng, Liu, Jin, Zeng, and Chen]{yu2023inpaint}
Tao Yu, Runseng Feng, Ruoyu Feng, Jinming Liu, Xin Jin, Wenjun Zeng, and Zhibo Chen.
\newblock Inpaint anything: Segment anything meets image inpainting.
\newblock \emph{arXiv preprint arXiv:2304.06790}, 2023.

\bibitem[Zhang et~al.(2017)Zhang, Xu, Li, Zhang, Wang, Huang, and Metaxas]{zhang2017stackgan}
Han Zhang, Tao Xu, Hongsheng Li, Shaoting Zhang, Xiaogang Wang, Xiaolei Huang, and Dimitris~N Metaxas.
\newblock Stackgan: Text to photo-realistic image synthesis with stacked generative adversarial networks.
\newblock In \emph{Proceedings of the IEEE international conference on computer vision}, pages 5907--5915, 2017.

\bibitem[Zhang et~al.(2024)Zhang, Mo, Chen, Sun, and Su]{zhang2024magicbrush}
Kai Zhang, Lingbo Mo, Wenhu Chen, Huan Sun, and Yu Su.
\newblock Magicbrush: A manually annotated dataset for instruction-guided image editing.
\newblock \emph{Advances in Neural Information Processing Systems}, 36, 2024.

\bibitem[Zhang et~al.(2023)Zhang, Yang, Feng, Qin, Chen, Yu, Chen, Wang, Savarese, Ermon, et~al.]{zhang2023hive}
Shu Zhang, Xinyi Yang, Yihao Feng, Can Qin, Chia-Chih Chen, Ning Yu, Zeyuan Chen, Huan Wang, Silvio Savarese, Stefano Ermon, et~al.
\newblock Hive: Harnessing human feedback for instructional visual editing.
\newblock \emph{arXiv preprint arXiv:2303.09618}, 2023.

\bibitem[Zhang et~al.(2021)Zhang, Tseng, Jiang, Yang, Lee, and Essa]{zhang2021text}
Tianhao Zhang, Hung-Yu Tseng, Lu Jiang, Weilong Yang, Honglak Lee, and Irfan Essa.
\newblock Text as neural operator: Image manipulation by text instruction.
\newblock In \emph{Proceedings of the 29th ACM International Conference on Multimedia}, pages 1893--1902, 2021.

\bibitem[Ziegler et~al.(2019)Ziegler, Stiennon, Wu, Brown, Radford, Amodei, Christiano, and Irving]{ziegler2019fine}
Daniel~M Ziegler, Nisan Stiennon, Jeffrey Wu, Tom~B Brown, Alec Radford, Dario Amodei, Paul Christiano, and Geoffrey Irving.
\newblock Fine-tuning language models from human preferences.
\newblock \emph{arXiv preprint arXiv:1909.08593}, 2019.

\end{thebibliography}
}

\setcounter{page}{0}
\renewcommand{\thefigure}{S\arabic{figure}}

\renewcommand{\thetable}{S\arabic{table}}
\renewcommand{\thesection}{\Alph{section}}
\setcounter{section}{0}  
\maketitlesupplementary

\section{Additional Model Outputs}

\begin{figure*}[b]
    \vspace{-0.1cm}
    \captionsetup{type=figure}
    \includegraphics[width=1\textwidth]{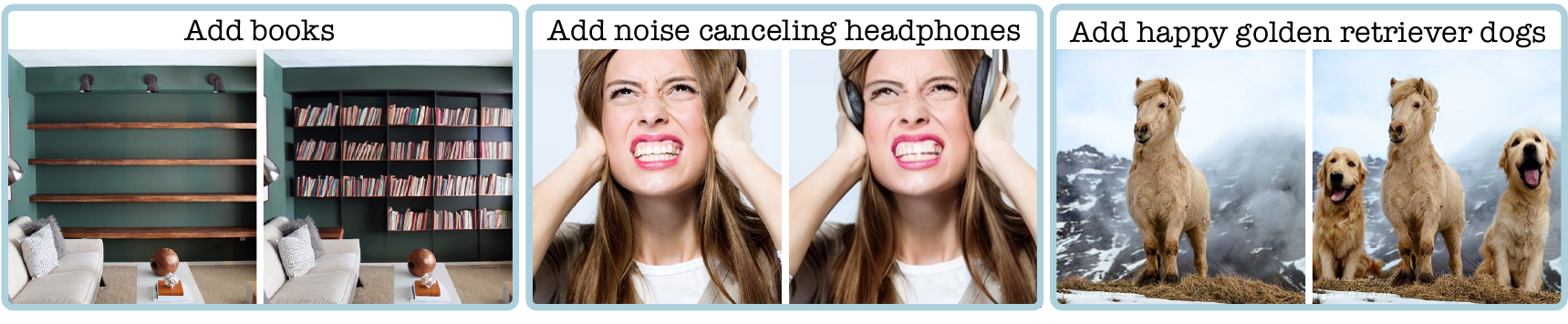}
    \includegraphics[width=1\textwidth]{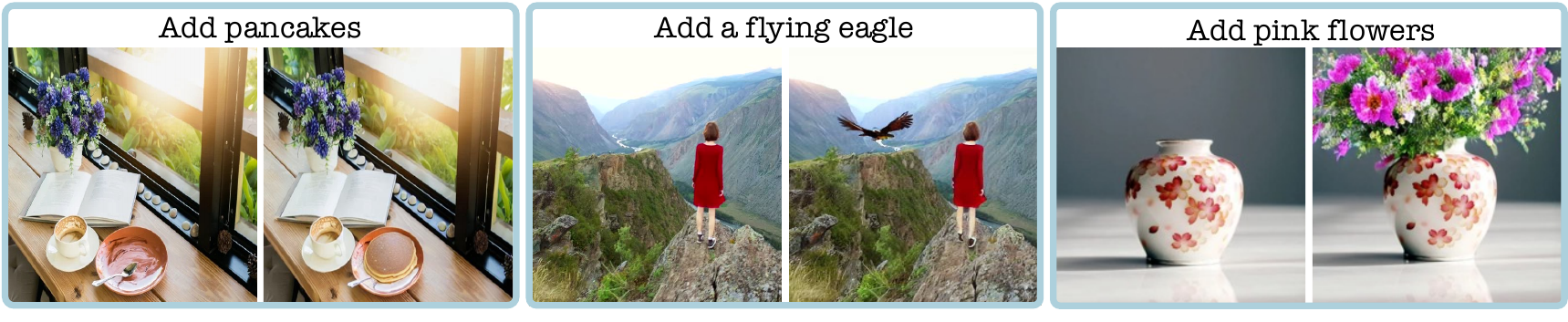}
    \includegraphics[width=1\textwidth]{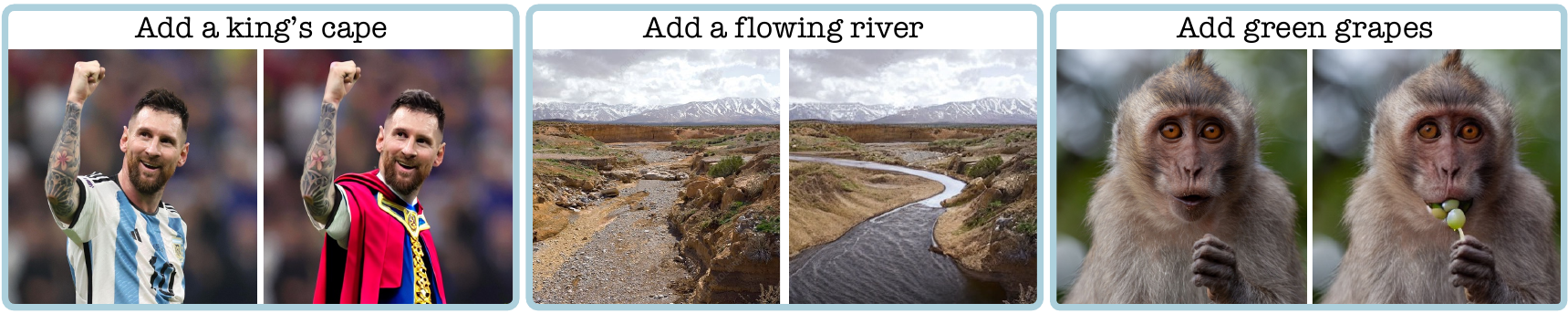}
    \vspace{-0.7cm}
    \caption{
    \textbf{Additional Object Addition Results of the Proposed Model.}
    The first two rows showcase outcomes from the model trained only with the \pipe{} dataset.
    The last row presents results from the same model after fine-tuning on the MagicBrush training set, as detailed in \Cref{sec:res1}.
    }
  \label{fig:supp_model_results}
\end{figure*}

\begin{figure}[t]
  \centering
  \vspace{-0.4cm}
  \includegraphics[width=0.8\linewidth]{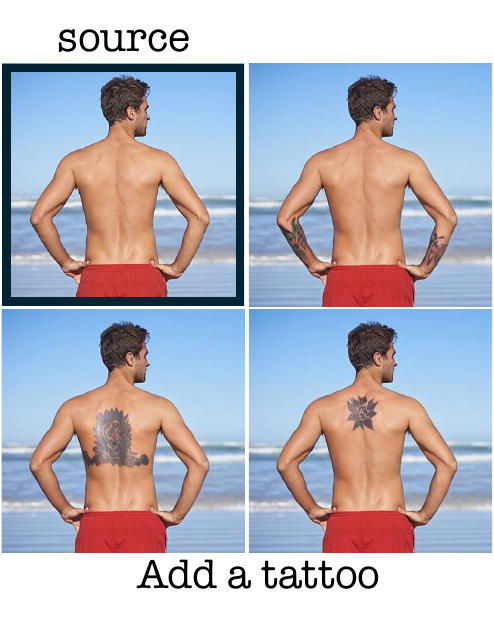}
    \vspace{-0.4cm}
    \caption{\textbf{Editing Diversity.} We generated three distinct edited images from the same source image, demonstrating the diversity of our model's outputs.}
   \vspace{-0.3cm}
  \label{Fig:Diversity}
\end{figure}

In continuation of the demonstrations seen in \Cref{Fig:Visual_Result_Teaser}, we further show a variety of object additions performed by our model in \Cref{fig:supp_model_results}.
The editing results showcase the model's ability to not only add a diverse assortment of objects and object types but also to integrate them seamlessly into images, ensuring the images remain natural and appealing.
Additionally, in \Cref{Fig:Diversity}, we provide an example of our model’s capability to generate diverse results for the same edit using different seeds.

\section{General Editing}
\begin{figure*}[h]
  \centering
  \includegraphics[width=0.95\linewidth]{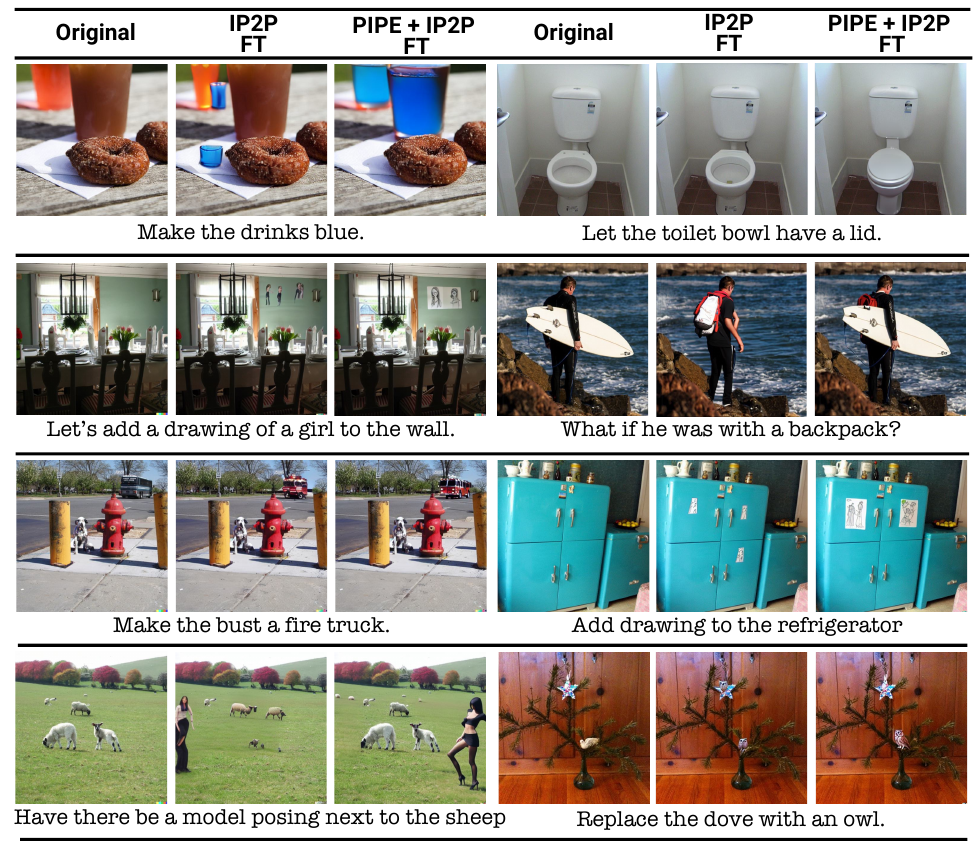}
   \vspace{-0.3cm}
  \caption{
  \textbf{Visual Comparison on General Editing Tasks.}
  The contribution of the \pipe{} dataset when combined with the IP2P dataset for general editing tasks, as evaluated on the full MagicBrush test set.
  The comparison is between a model trained on these merged datasets and a model trained solely on the IP2P dataset, with both models fine-tuned on the MagicBrush training set.
  The results demonstrate that, although the \pipe{} dataset focuses solely on object addition instructions, it enhances performance across a variety of editing tasks.
  \\ \\ \\ \\
  }
   \vspace{-0.3cm}
  \label{Fig:Visual_Full_Brush}
\end{figure*}

As detailed in \Cref{section:ip2p_merge}, the model, trained on the combined IP2P and \pipe{} dataset, achieves new state-of-the-art scores for the general editing task.
In \Cref{Fig:Visual_Full_Brush}, we present a visual comparison that contrasts our model's performance with that of a model trained without the \pipe{} dataset. The results not only underscore our model's superiority in object additions but also demonstrate its effectiveness in enhancing outcomes for other complex tasks, such as object replacement.

We further analyze this model by testing its performance not on the entire MagicBrush dataset as in \Cref{section:ip2p_merge}, but on the 'addition only' subset (discussed in \Cref{supp:subset}) and its complementary 'not addition' subset.
The experiments are performed under the same configuration as \Cref{section:ip2p_merge}.
Results for the addition subset and the complementary subset are presented in \Cref{Table:SP_add_noadd_Brush}.
In both subsets, our model outperforms the other models, indicating that although our dataset focuses on adding instructions, the inclusion of a large amount of high-quality editing data enhances performance for general editing tasks as well.

\vspace{-0.1cm}
\section{Multiple Object Addition}\label{sec:multiple_objects}
\begin{figure}[t]
  \centering
  \includegraphics[width=1\linewidth]{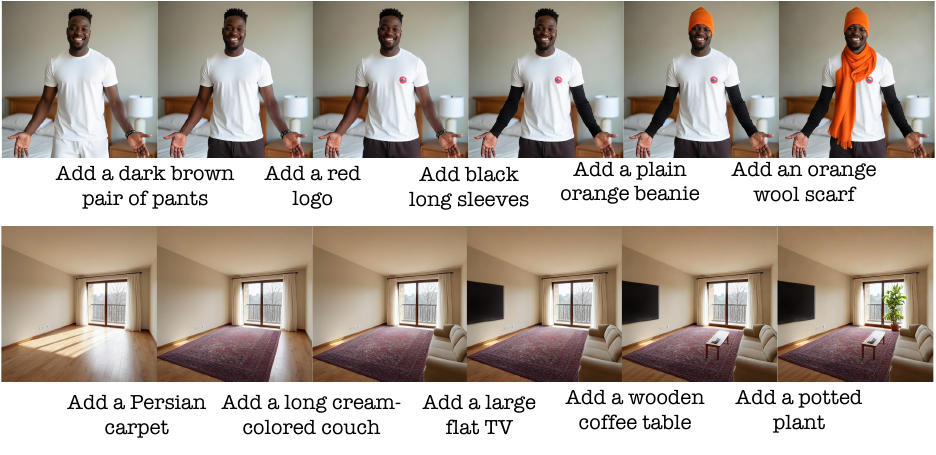}
   \caption{
   \textbf{Multiple objects.} 
   Example of multiple object addition using latent editing, where successive objects are added without intermediate decoding and encoding.}
   \vspace{-0.3cm}
   \label{fig:multiple_objects}
\end{figure}

\vspace{-0.1cm}
A straightforward extension of our model allows for adding multiple objects by applying it recurrently, each time with a different addition prompt.
However, this approach poses a challenge because each object addition requires a decode–encode cycle through the Stable Diffusion variational autoencoder (VAE), where each pass degrades image quality (see \cref{SM_figure:consistency_enforcment}).
To mitigate this, we perform all edits in latent space—encoding only before the first addition and decoding only after the final addition.
\cref{fig:multiple_objects} illustrates this process, demonstrating the successful addition of objects without intermediate decoding.

\begin{table*}[t]
  \centering
  \setlength\tabcolsep{0pt}
  \begin{tabular*}{\textwidth}
    {@{\extracolsep{\fill}}l
    @{\hspace{0.4em}} c
    @{\hspace{0.6em}} c
    @{\hspace{0.2em}} c
    @{\hspace{0.2em}} c
    @{\hspace{0.2em}} c
    |
    @{\hspace{0.4em}} c
    @{\hspace{0.6em}} c
    @{\hspace{0.2em}} c
    @{\hspace{0.2em}} c
    @{\hspace{0.2em}} c
    }
    \toprule
    \multicolumn{6}{c|}{\textbf{Addition Subset}} & \multicolumn{5}{c}{\textbf{Non-Addition Subset}} \\
    \midrule
    \textbf{Methods} & L1$_{\downarrow}$ & L2$_{\downarrow}$ & CLIP-I$_{\uparrow}$ & DINO$_{\uparrow}$ & CLIP-T$_{\uparrow}$ & L1$_{\downarrow}$ & L2$_{\downarrow}$ & CLIP-I$_{\uparrow}$ & DINO$_{\uparrow}$ & CLIP-T$_{\uparrow}$ \\
    \midrule
    IP2P & .100 & .031 & .860 & .700 & .289 & .114 & .038 & .839 & .742 & .290 \\ 
    IP2P FT  & .077 & .028 & .902 & .867 & .306 & .083 & .032 & .895 & .841 & .300 \\
    Ours + IP2P FT & \textbf{.069} & \textbf{.024} & \textbf{.913} & \textbf{.889} & \textbf{.308} & \textbf{.075} & \textbf{.027} & \textbf{.905} & \textbf{.862} & \textbf{.303} \\
    \bottomrule 
  \end{tabular*}
  \caption{\textbf{Global Editing Performance on Addition and Non-Addition MagicBrush Subsets.}
    Evaluation of our global editing model performance on both the add and complementary non-add instruction subsets of MagicBrush.
    The model, trained on the combined PIPE and IP2P datasets and fine-tuned on the MagicBrush training set, surpasses IP2P and the fine-tuned IP2P models in both subsets.
    }
    \vspace{-0.3cm}
  \label{Table:SP_add_noadd_Brush}
\end{table*}

\section{Limitations}
\label{sec:limitations}
\begin{figure*}[h]
  \centering
  \includegraphics[width=1\linewidth]{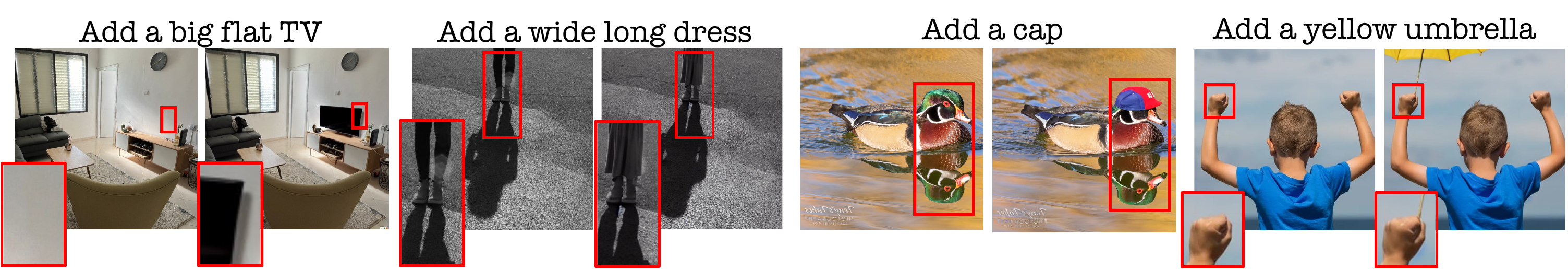}
   \vspace{-0.15cm}
  \caption{
  \textbf{Limitations.}
    Left: Successful shadow generation near the object. Center: Failures in generating shadows or reflections when distant from the object. Right: Failure in changing hand posture and maintaining the original one.
    }
  \label{SM_figure:limitations}
\end{figure*}

Despite the impressive results produced by our model, several limitations remain. First, while our data curation pipeline improves robustness during the removal phase, it is not entirely error-free. Additionally, the model struggles with significant changes occurring far from the object but are affected by it. For instance, it handles nearby effects, like TV shadows (see \cref{SM_figure:limitations}), but struggles with larger shadows or distant reflections, as seen in the center images of \cref{SM_figure:limitations}. Similarly, object-object interactions are not always accurately handled (see the right images in the figure). These challenges stem from the dataset construction, as our method minimizes alterations outside the near-object region. Future work could explore inpainting both the object and distant regions influenced by it. We hope our work inspires future research to address these limitations.

\begin{figure*}[h]
  \centering
  \includegraphics[width=\linewidth]{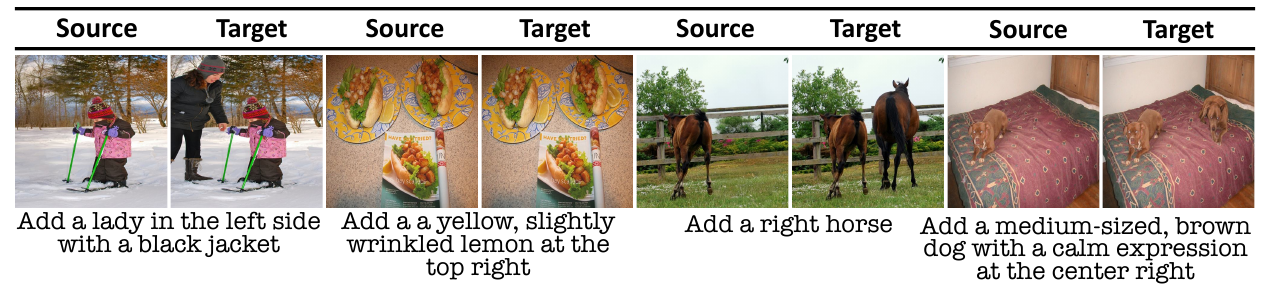}
  \vspace{-0.3cm}
  \includegraphics[width=\linewidth]{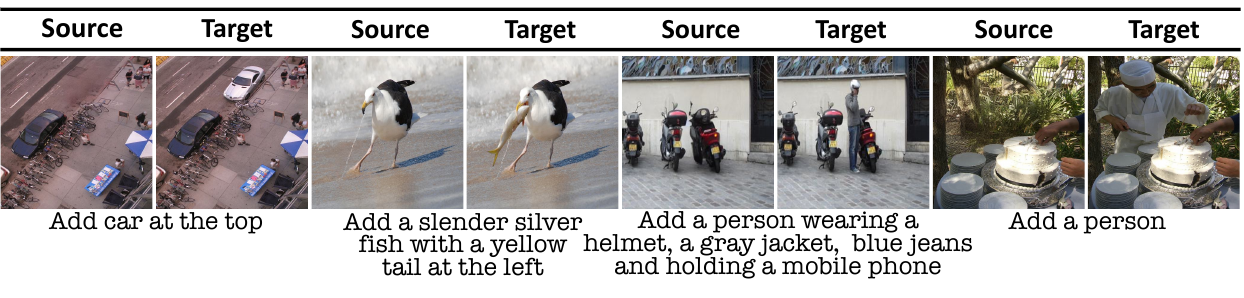}
  \caption{\textbf{Additional \pipe{} Datasets Examples.}
  }
  \label{Fig:supp_Dataset_Examples}
  \vspace{-0.3cm}
\end{figure*}

\section{PIPE Dataset}\label{supp:data}

\subsection{Creating Source-Target Image Pairs}
We offer additional details on the post-removal steps described in \Cref{sec:method:dataset}.
The post-removal process involves assessing the CLIP similarity between the class name of the removed object and the inpainted area.
This assessment helps evaluate the quality of the object removal, ensuring no objects from the same class remain.
To measure CLIP similarity for the inpainted area only, we counter the challenge of CLIP's unfamiliarity with masked images by reducing the background's influence on the analysis. We do this by adjusting the background to match the image's average color and integrating the masked area with this unified background color.
A dilated mask smoothed with a Gaussian blur is employed to soften the edges, facilitating a more seamless and natural-looking blend.

To complement the CLIP score similarity, we introduce an additional measure that quantifies the shift in similarity before and after removal.
Removals with a high pre-removal similarity score, followed by a comparatively lower yet significant post-removal score are not filtered, even though they exceed the threshold.
This method allows for the efficient exclusion of removals, even when other objects of the same class are in close spatial proximity.

\begin{figure}[t]
    \centering
    \centering
        \includegraphics[width=\linewidth]{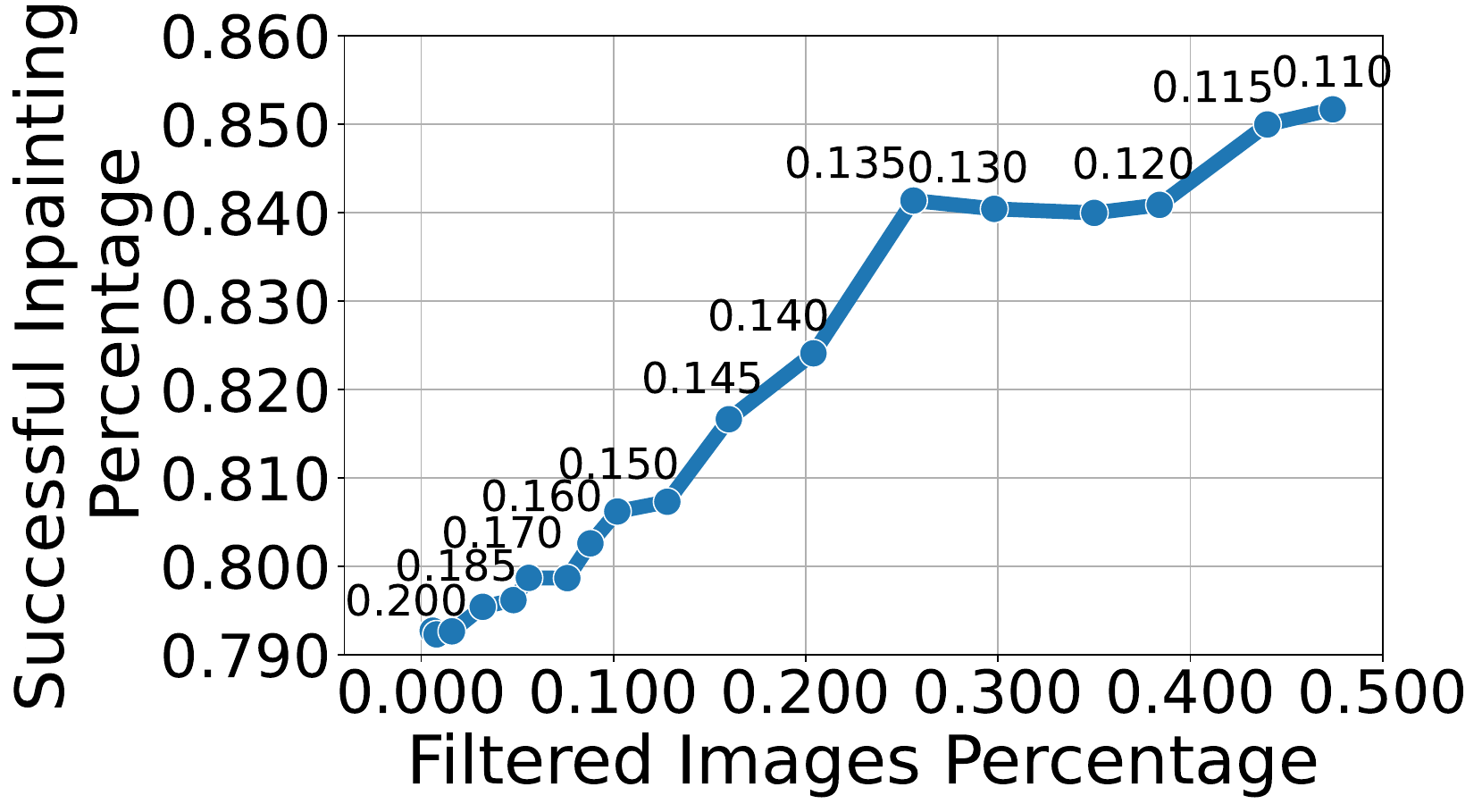}
        \captionsetup{skip=2pt} 
        \caption{
        \textbf{Concensus Filtering Success for varying Thresholds}
        }
        \vspace{-0.5cm}
        \label{SM_figure:Concensus_treshold}
\end{figure}

\begin{figure}[t]
        \centering
        \includegraphics[width=\linewidth]{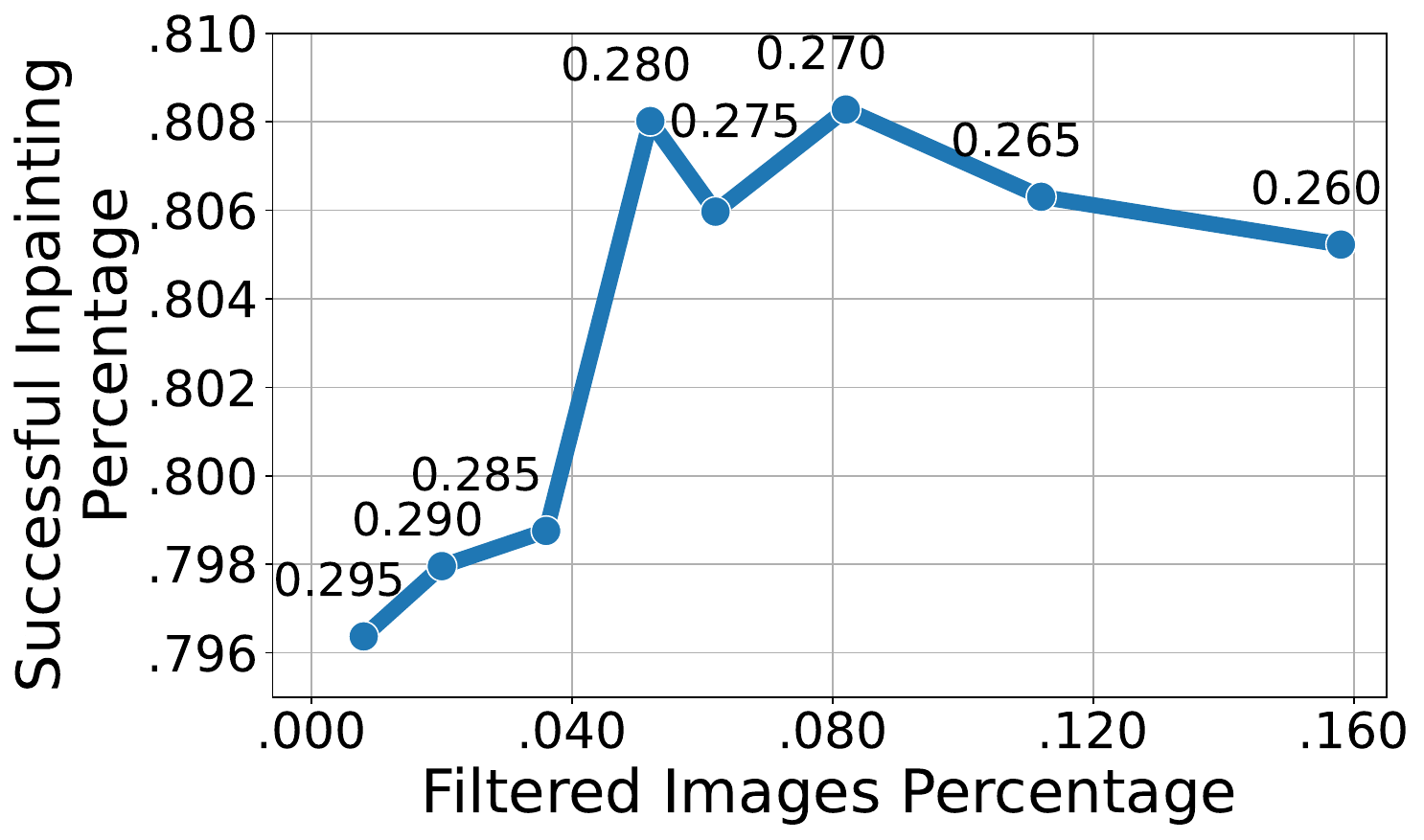}
        \captionsetup{skip=2pt} 
        \caption{
        \textbf{Multimodal CLIP Filtering Success for varying Thresholds}}
        \vspace{-0.5cm}
        \label{SM_figure:Multimodal_treshold}
\end{figure}

\Cref{SM_figure:Concensus_treshold} and \Cref{SM_figure:Multimodal_treshold} present the figures discussed in \Cref{sec:method:dataset} related to filtering thresholds and their justification.
\Cref{Table:datasets_filter_stats} reports the number of images before and after each filtering stage.

\begin{figure*}[!htbp]
    \centering
    \begin{minipage}[t]{0.56\textwidth}
        \centering
        \includegraphics[width=\textwidth]{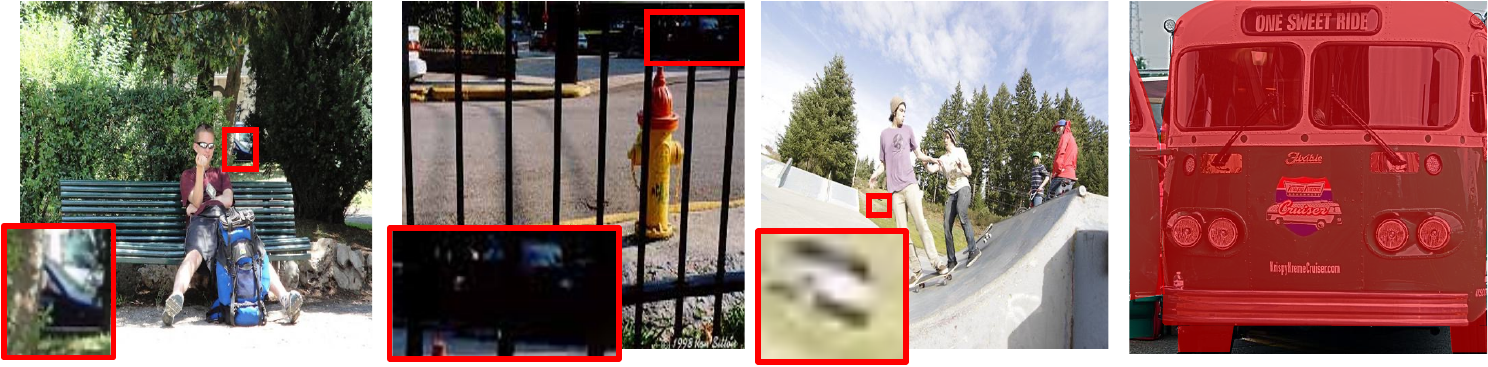}
        \captionsetup{skip=2pt} 
        \caption{
        \textbf{Pre-Removal Filtered Examples.}
        Left: Objects with non-informative view and low CLIP Object similarity.
        Right: Extremely small and large objects, unsuitable for our dataset.
        }
        \label{SM_figure:pre_removal}
    \end{minipage}
    \hfill
    \begin{minipage}[t]{0.42\textwidth}
        \centering
        \includegraphics[width=\textwidth]{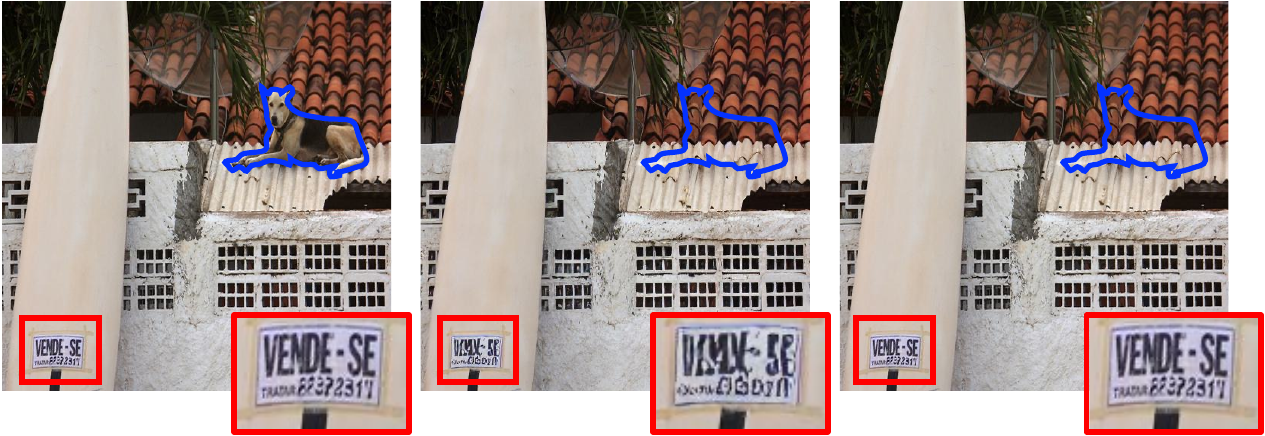}
        \captionsetup{skip=2pt} 
        \caption{
        \textbf{Consistency Enforcement Examples.}
        From left to right: original image, inpainted dog image, inpainted image after alpha blending.
        }
        \label{SM_figure:consistency_enforcment}
    \end{minipage}
\end{figure*}

\begin{table}[h]
\caption{\textbf{Statistics on the dataset before and after each Filtering step.}}
\vspace{-0.25cm}
\small
\begin{tabular*}{1.03\linewidth}{@{\extracolsep{\fill}}cccccc} 
\toprule
{ Initial} & {Pre-Removal} & {Consensus} & {MM CLIP} & {{Importance}}\\
\midrule
4,646K & 1,494K & 1,101K & 986K & 888K \\
\bottomrule 
\end{tabular*}
\label{Table:datasets_filter_stats}
\vspace{-0.4cm}
\end{table}

\subsection{VLM-LLM Based Instructions}
Using a VLM and an LLM, we convert the class names of objects from the segmentation dataset into detailed natural language instructions (\Cref{section:instruction_generation}).
Initially, for each image, we present the masked image (featuring only the object) to CogVLM with the prompt:
``\texttt{Accurately describe the main characteristics of the <class name>.
Use few words which best describe the <class- name>}''.
This process yields an in-depth description centered on the object, highlighting key attributes such as shape, color, and texture. Subsequently, this description is provided to the LLM along with human-crafted prompts for In-Context Learning (ICL), to generate succinct and clear instructions.
The implementation of the ICL mechanism is detailed in \Cref{tab:appendix-icl}.

Furthermore, we enrich the instructions by including a coarse language-based description of the object's location within the image, derived from the given mask. To accomplish this, we split the image into a nine-section grid and assign each section a descriptive label (e.g., top-right).
This spatial description is then randomly appended to the instruction with a $25\%$ probability during the training process.

\subsection{Integrating Instruction Types}
As detailed in \Cref{section:instruction_generation}, we construct our instructions using three approaches: (i) class name-based (ii) VLM-LLM based, and (iii) manual reference-based.
These three categories are then integrated to assemble the final dataset.
The dataset includes 887,773 instances each from Class name-based and VLM-LLM-based methods, with an additional 104,373 from Manual reference-based instructions.

\subsection{Additional Examples}
In \Cref{Fig:supp_Dataset_Examples}, we provide further instances of the \pipe{} dataset that complement those in \Cref{Fig:Dataset_Examples}.

\vspace{-0.1cm}

\begin{table*}[th]
\centering
\small
\caption{\textbf{In-Context Learning Prompt}. (Top) We provide the model with five examples of captions and their corresponding human-annotated responses. (Bottom) We introduce it with a new caption and request it to provide an instruction.}
\vspace{0.2cm}
\begin{tabular}{l}
\toprule
$[\text{\textcolor{blue}{\textbf{USER}}}]$: Convert the following sentence into a short image addition instruction: \\ <caption 0>. \\ Use straightforward language and describe only the <class name 0>. \\ Ignore surroundings and background and avoid pictorial description.\\
$[\text{\textcolor{purple}{\textbf{ASSISTANT}}}]$: <example response 0>
\\ 
$\vdots$
\\
$[\text{\textcolor{blue}{\textbf{USER}}}]$: Convert the following sentence into a short image addition instruction: \\ <caption 4>. \\ Use straightforward language and describe only the <class name 4>. \\ Ignore surroundings and background and avoid pictorial description.\\
$[\text{\textcolor{purple}{\textbf{ASSISTANT}}}]$: <example response 4> \\ \\
\cdashline{1-1}
\\
$[\text{\textcolor{blue}{\textbf{USER}}}]$: Convert the following sentence into a short image addition instruction: \\ <new caption>. \\ Use straightforward language and describe only the <new class name>. \\ Ignore surroundings and background and avoid pictorial description.\\
$[\text{\textcolor{purple}{\textbf{ASSISTANT}}}]$: 
\\ 
\bottomrule
\end{tabular}
\label{tab:appendix-icl}
\end{table*}


\section{Implementation Details}\label{supp:implementation}


As noted in \Cref{section:model_train}, the training of our editing model is initialized with the SD v1.5 model.
Conditions are set with $c_T = \varnothing$, $c_I = \varnothing$, and both $c_T =  c_I = \varnothing$ occurring with a $5\%$ probability each.
The input resolution during training is adjusted to 256, applying random cropping for variation.
Each GPU manages a batch size of 128.
The model undergoes training for $60$ epochs, utilizing the ADAM optimizer. It employs a learning rate of $5\cdot10^{-5}$, without a warm-up phase.
Gradient accumulation is set to occur over four steps preceding each update, and the maximum gradient norm is clipped at 1.
Utilizing eight NVIDIA A100 GPUs, the total effective batch size, considering the per-GPU batch size, the number of GPUs, and gradient accumulation steps, reaches $4096$ $(128 \cdot 8 \cdot 4)$.

For the fine-tuning phase on the MagicBrush training set (\Cref{sec:res1}), we adjust the learning rate to $10^{-6}$ and set the batch size to $8$ per GPU, omitting gradient accumulation, and train for $250$ epochs.

\vspace{-0.2cm}
\subsection{MagicBrush Subset}\label{supp:subset}
\vspace{-0.2cm}
To initially focus our analysis on the specific task of object addition, we applied an automated filtering process to the MagicBrush dataset.
This process aims to isolate image pairs and associated instructions that exclusively pertained to object addition.
To ensure an unbiased methodology, we applied an automatic filtering rule across the entire dataset.
The filtering criterion applied retained instructions explicitly containing the verbs "add" or "put," indicating object addition. Concurrently, instructions with "remove" were excluded to avoid object replacement scenarios, and those with the conjunction "and" were omitted to prevent cases involving multiple instructions.

\subsection{Evaluation} 
In our comparative analysis in \Cref{sec:res1}, we assess our model against leading instruction-following image editing models.
To ensure a fair and consistent evaluation across all models, we employed a fixed seed ($0$) for all comparisons.

Our primary analysis focuses on two instruction-guided models,  IP2P~\citep{brooks2023instructpix2pix} and Hive~\citep{zhang2023hive}.
For IP2P, we utilized the Hugging Face diffusers model and pipeline\footnote{\url{https://hf.co/docs/diffusers/training/instructpix2pix}}, adhering to the default inference parameters.
Similarly, for Hive, we employed the official implementation provided by the authors\footnote{\url{https://github.com/salesforce/HIVE}}, with the documented default parameters.

\begin{figure*}[t]
  \centering
  \includegraphics[width=\linewidth]{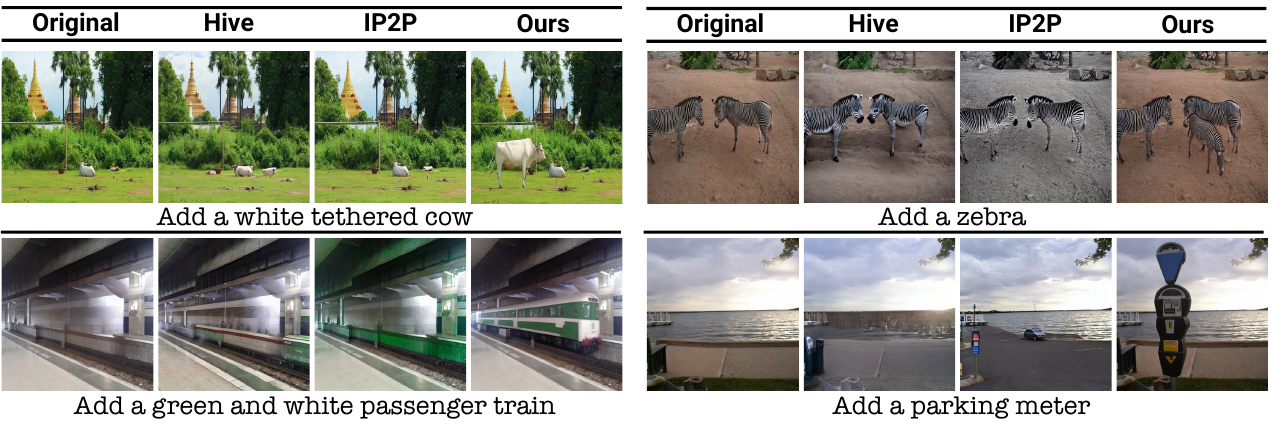}
\vspace{-0.6cm}
  \caption{\textbf{ Visual Comparison of the Proposed Model on \pipe{} Test Set.}
  The visual evaluation highlights the effectiveness of our method against other leading models on the \pipe{} test set. 
  Our model excels in adhering closely to specified instructions and accurately generating objects in terms such as style, scale, and location.
  }
  \label{Fig:supp_Visual_Comparison_PIPE}
\end{figure*}

\begin{figure*}[t]
  \centering
  \includegraphics[width=\linewidth]{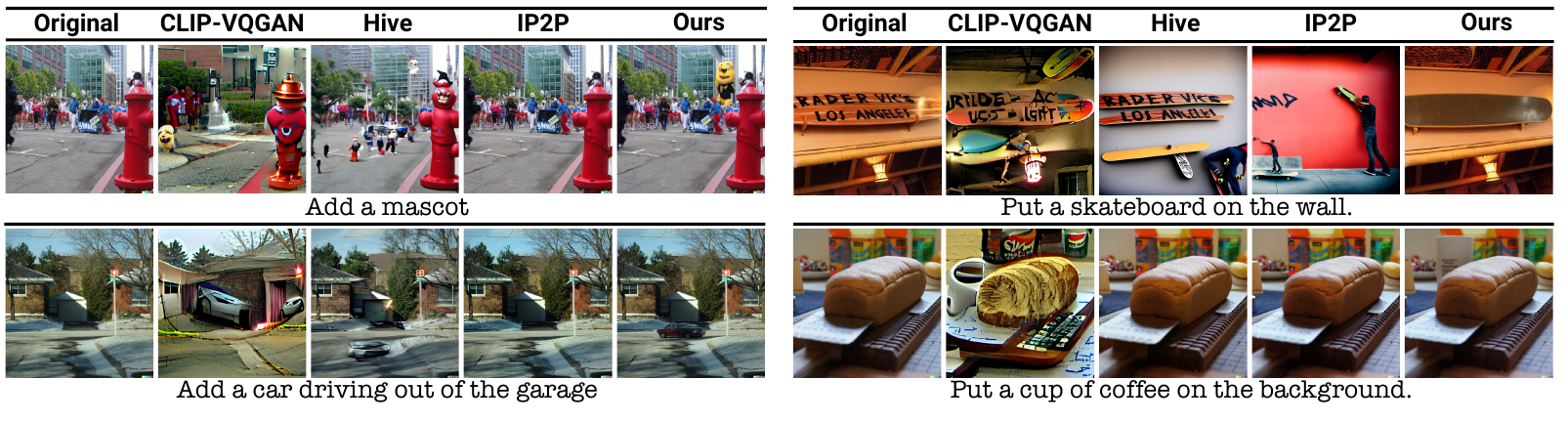}
  \vspace{-0.6cm}
  \caption{\textbf{Visual Comparison of the Proposed Model on MagicBrush Test Subset.} 
  Our method versus leading models within the MagicBrush object addition test subset.
  It illustrates our model's superior generalization across varied instructions and datasets, outperforming the other approaches.
  }
  \label{Fig:supp_Visual_Comparison_BRUSH}
\end{figure*}

\begin{figure*}[!h]
  \centering
  \begin{minipage}{0.47\textwidth}
    \centering
    \includegraphics[width=\textwidth]{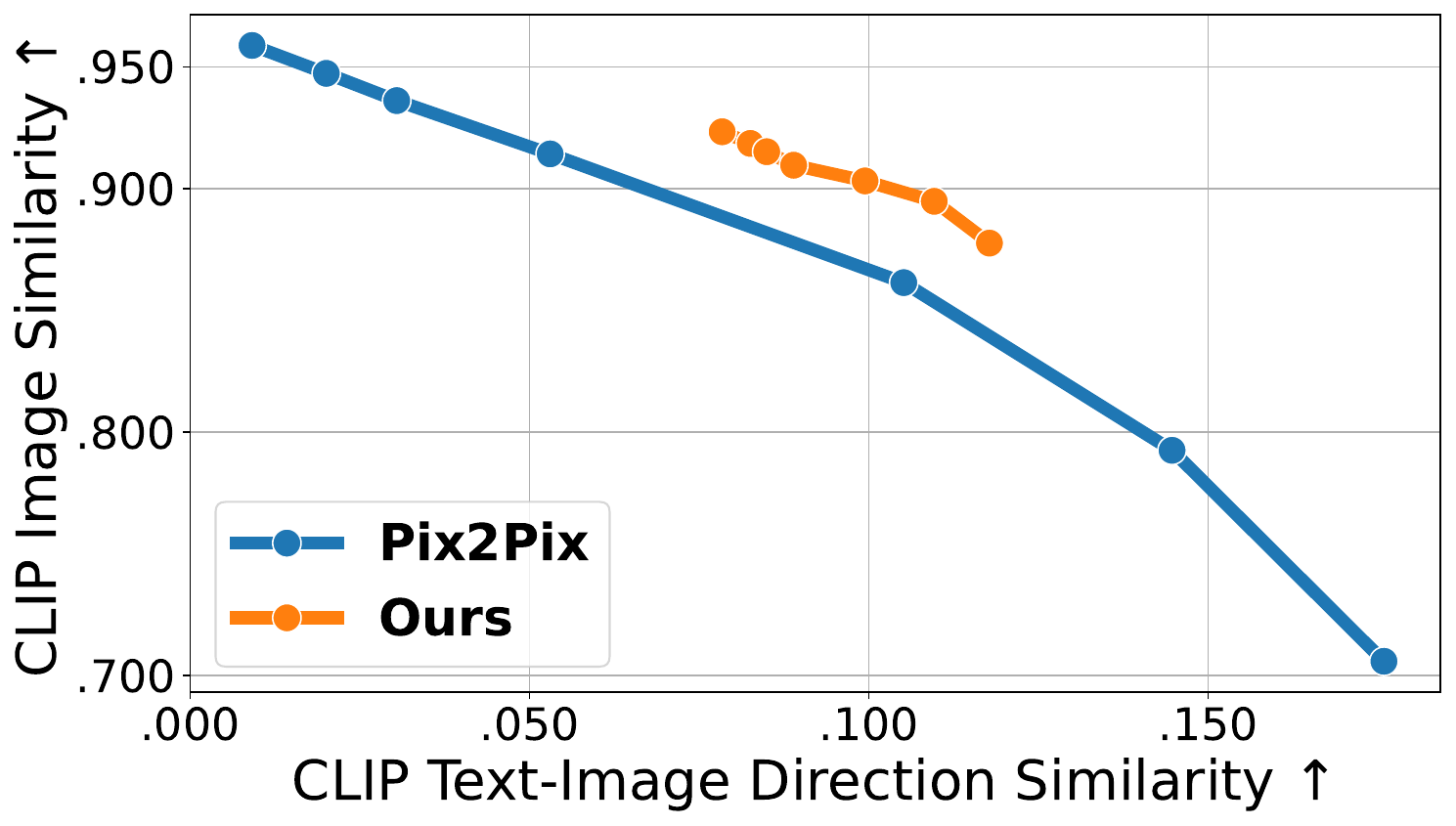}
    \caption{
    \textbf{Model Consistency-Instruction Trade-off:} Trade-off between consistency with the input image (Y-axis) and edit adherence (X-axis) for IP2P and our model on the MagicBrush test subset. Text guidance is fixed at $7$, and image guidance ranges from $1$ to $2.5$.
    \\
    }
    \label{fig:brush_add_plot}
  \end{minipage}
  \hfill
  \begin{minipage}{0.49\textwidth}
    \centering
    \includegraphics[width=0.96\textwidth]{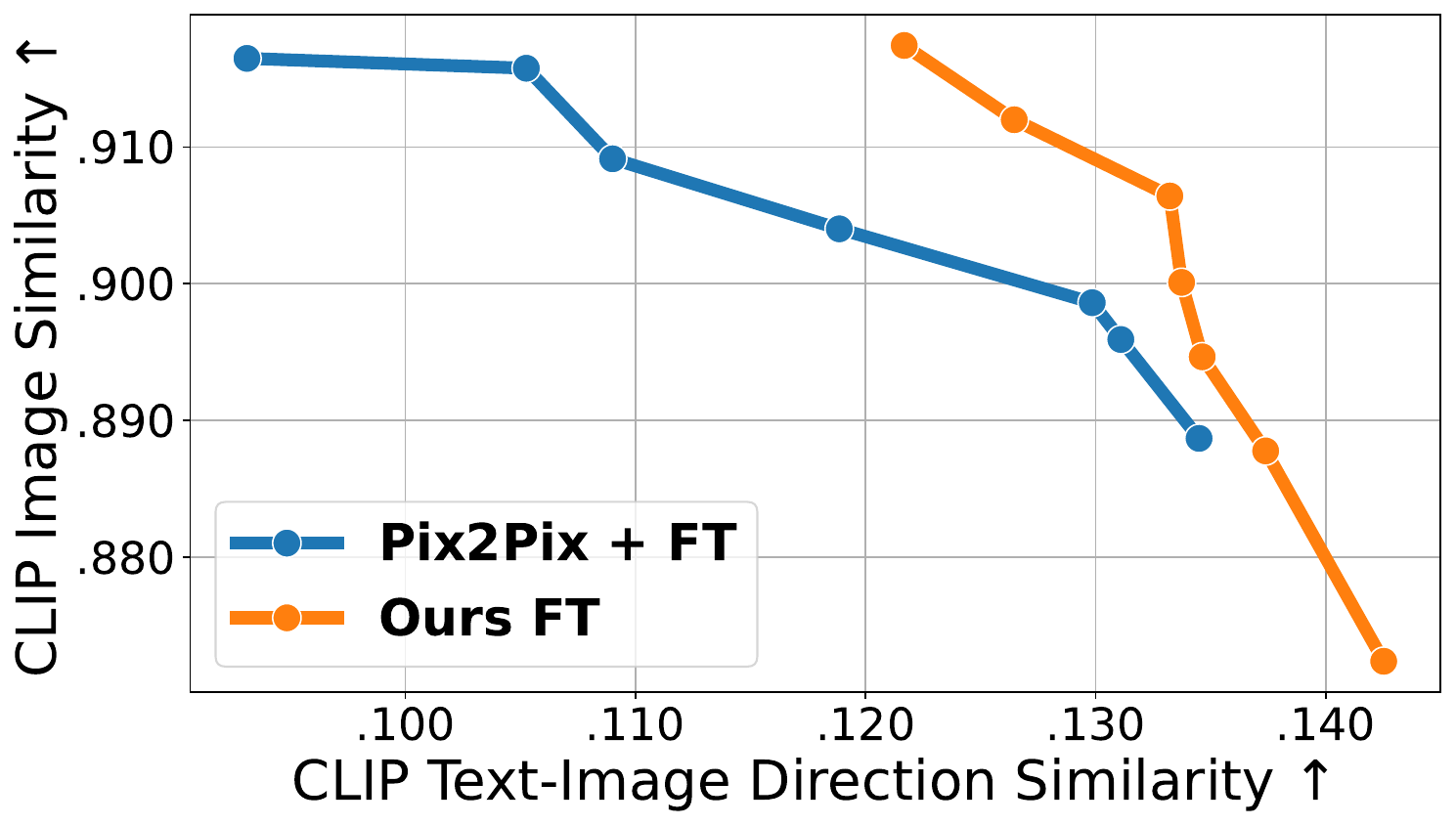}
    \caption{
    \textbf{Finetuned-Model Consistency-Instruction Trade-off:} Trade-off between consistency with the input image (Y-axis) and edit adherence (X-axis) for IP2P and our model, both fine-tuned on the MagicBrush training set and tested on its test subset. Text guidance is fixed at $7$, and image guidance ranges from $1$ to $2.5$.
    }
    \label{fig:brush_add_FT_plot}
  \end{minipage}
\end{figure*}

 Our comparison extends to models that utilize global descriptions: VQGAN-CLIP~\citep{crowson2022vqganclip} Null-Text-Inversion~\citep{mokady2023null}, Pix2PixZero~\citep{parmar2023zero}, Edit-Freindly DDPM~\citep{huberman2024edit} and SDEdit~\citep{meng2021sdedit}.
These models were chosen for evaluation within the MagicBrush dataset, as global descriptions are not available in both the OPA and our PIPE dataset.
For VQGAN-CLIP\footnote{\url{https://github.com/nerdyrodent/VQGAN-CLIP}}, Null-Text-Inversion\footnote{\url{https://github.com/google/prompt-to-prompt/blob/main/null_text_w_ptp.ipynb}} and Edit-Freindly DDPM\footnote{\url{https://github.com/inbarhub/DDPM_inversion}}, we used the official code base with the default hyperparameters. For SDEdit\footnote{\url{https://hf.co/docs/diffusers/en/api/pipelines/stable_diffusion/img2img}} and Pix2PixZero\footnote{\url{https://hf.co/docs/diffusers/main/en/api/pipelines/pix2pix_zero}}, we used the image-to-image pipeline of the Diffusers library with the default parameters.

\begin{table*}[t]
    \centering
    \normalsize 
    \begin{tabular}{lcccccccc}
        \toprule
        \textbf{Model}  & VQGAN-CLIP & SDEdit & NTI & P2P-Z & EFD & Hive & IP2P & Ours \\
        \midrule
        \textbf{VQAScore} & 0.7675 & 0.6114 & 0.6008 & 0.5356 & 0.6792 & 0.5822 & 0.5408 & 0.7045 \\
        \bottomrule
    \end{tabular}
    \vspace{-0.2cm}
    \caption{\textbf{VQAScore Metric.} We use VQAScore~\cite{lin2024evaluating} as a VQA-based alignment metric to further evaluate our method.}
    \label{Table:vqa_scores}
    \vspace{-12pt}
\end{table*}

\begin{figure}[t]
  \centering
  \includegraphics[width=0.8\linewidth]{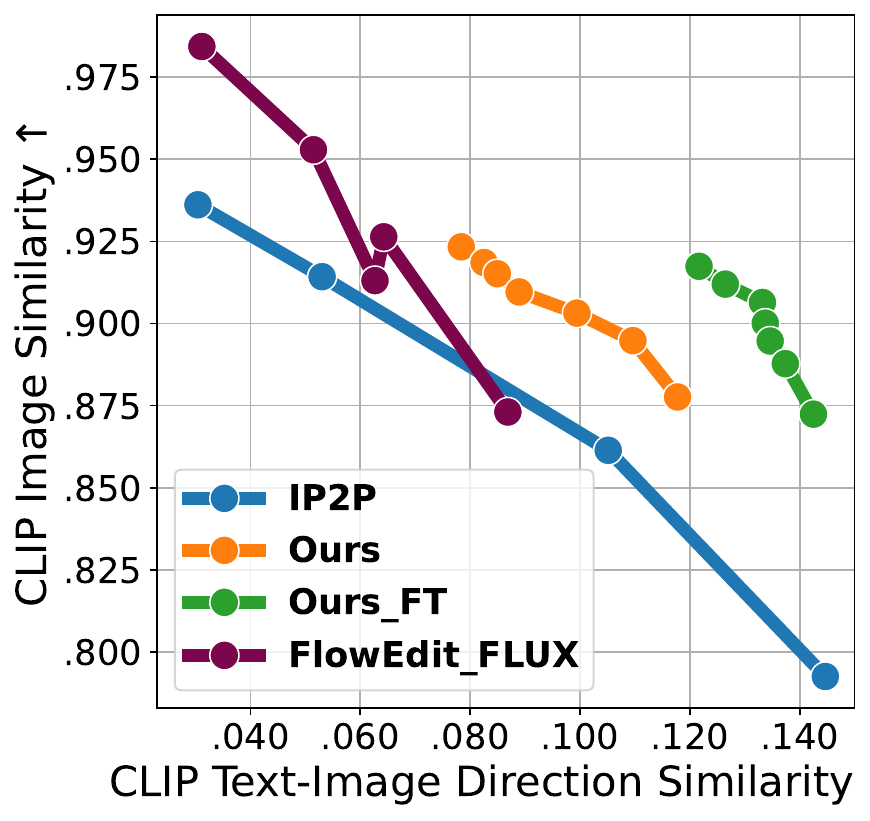}
  \vspace{-0.2cm}
  \caption{\textbf{Comparison to FLUX-Based Method.} 
  We compared our model to a FLUX-based approach, specifically the recently released FlowEdit~\cite{kulikov2024flowedit}. As shown in the plot, our model significantly outperforms FlowEdit. }
  \label{Fig:flux_comparison}
  \vspace{-0.1cm}
\end{figure}

We also evaluated our fine-tuned model against the MagicBrush fine-tuned model, as documented in ~\citep{zhang2024magicbrush}.
Although this model does not serve as a measure of generalizability, it provides a valuable benchmark within the specific context of the MagicBrush dataset.
For this comparison, we employed the model checkpoint and parameters as recommended on the official GitHub repository of the MagicBrush project\footnote{\url{https://github.com/OSU-NLP-Group/MagicBrush}}. 
In~\Cref{Fig:supp_Visual_Comparison_PIPE} and~\Cref{Fig:supp_Visual_Comparison_BRUSH}, we provide additional qualitative examples on the tested datasets to complement the ones in \Cref{Fig:Visual_Comparison}.
We further assess the model's performance on the MagicBrush subset using the same CLIP Image similarity versus Directional CLIP similarity measure, as explained in \Cref{section:ip2p_merge}.
We plot this measure to compare the IP2P model with our model in \Cref{fig:brush_add_plot} and the MagicBrush fine-tuned models in \Cref{fig:brush_add_FT_plot}.
As shown in both comparisons, our models present a better trade-off between consistency with the input image and adherence to the edit instruction, achieving higher consistency with the instruction for the same similarity to the input image.

\begin{figure*}[t]
  \centering
  \includegraphics[width=\linewidth]{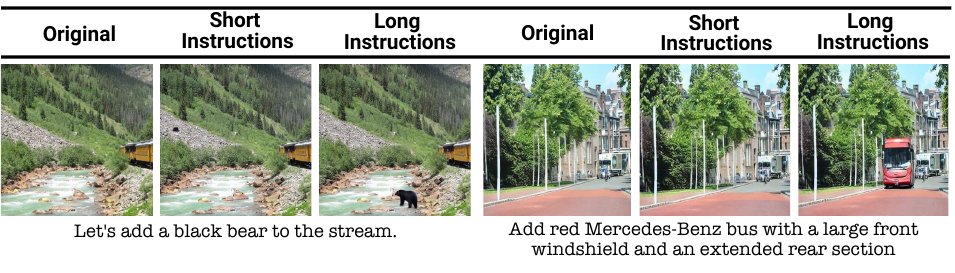}
  \vspace{-0.8cm}
  \caption{\textbf{Instructions Ablation Examples.}
    Qualitative comparison of model performance when trained on 'short' template-based instructions versus 'long' instructions generated through our VLM-LLM pipeline.
    Models trained on the latter exhibit superior performance in interpreting complex instructions and closely aligning object additions with editing requests.
    } 
  \label{Fig:ablation}
    \vspace{-0.1cm}
\end{figure*}

To further evaluate our method with more advanced metrics, we extend \Cref{Table:MagicBrush} and report the VQAScore \cite{lin2024evaluating}, as shown in \Cref{Table:vqa_scores}.
Under this metric, our approach maintains its favorable performance, except for VQGAN-CLIP, which, as discussed in \Cref{pargagrh:mb}, tends to deviate substantially from the original image.
Furthermore, we extend \Cref{fig:comp_real} by adding a comparison between our model and a FLUX-based \cite{Flux2024} approach, the recently released FlowEdit \cite{kulikov2024flowedit}, demonstrating superior performance.




\begin{table*}[t]
\setlength\tabcolsep{0pt}
\centering
\begin{tabular*}{1\textwidth}{@{\extracolsep{\fill}}ccccccc} 
\toprule
{\textbf{   Train Instructions Type   }} & { L1 $\downarrow$} & {L2 $\downarrow$} & {CLIP-I $\uparrow$} & {DINO$\uparrow$} & {CLIP-T $\uparrow$}\\
\midrule
Short Instructions  & 0.083 & 0.028  & \textbf{0.900} & \textbf{0.856}& 0.300 \\ 
Long Instructions & \textbf{0.072} & \textbf{0.025} & \textbf{0.900} & 0.852 & \textbf{0.302}\\
\bottomrule 
\end{tabular*}
\vspace{-0.2cm}
\caption{\textbf{Instructions Ablation Analysis.}
A quantitative comparative analysis of model performance, comparing training on 'short' class-based instructions to 'long' instructions generated using the VLM and LLM pipeline. 
This analysis was performed on MagicBrush subset.
The results demonstrate that training with VLM-LLM-based instructions significantly enhances performance, thereby confirming its effectiveness.}
\vspace{-0.3cm}
\label{Table:ablation}
\end{table*}

\section{Instructions Ablation}
We examine the impact of employing our VLM-LLM pipeline, detailed in \Cref{section:instruction_generation}, for generating natural language instructions.
The outcomes of the pipeline, termed "long instructions", are compared with brief, class name-based instructions (\textit{e.g.}, ``\texttt{Add a cat}''), referred to as "short instructions".
In \Cref{Table:ablation}, we assess a model trained on the \pipe{} image pairs, comparing its performance when trained with either long or short inputs.
The models are evaluated on MagicBrush subset. As expected, training with long instructions leads to improved performance on MagicBrush. This demonstrates that training with comprehensive instructions generated by our VLM-LLM mechanism benefits at inference time.
In addition to quantitative results, we provide qualitative results of both models in \Cref{Fig:ablation}.
As illustrated, the model trained with long instructions shows superior performance in interpreting complex instructions that include detailed descriptions and location references, such as "Let's add a black bear to the stream".

\section{Human Evaluation}
\begin{figure*}[!htbp]
    \centering
    \includegraphics[width=0.78\textwidth]{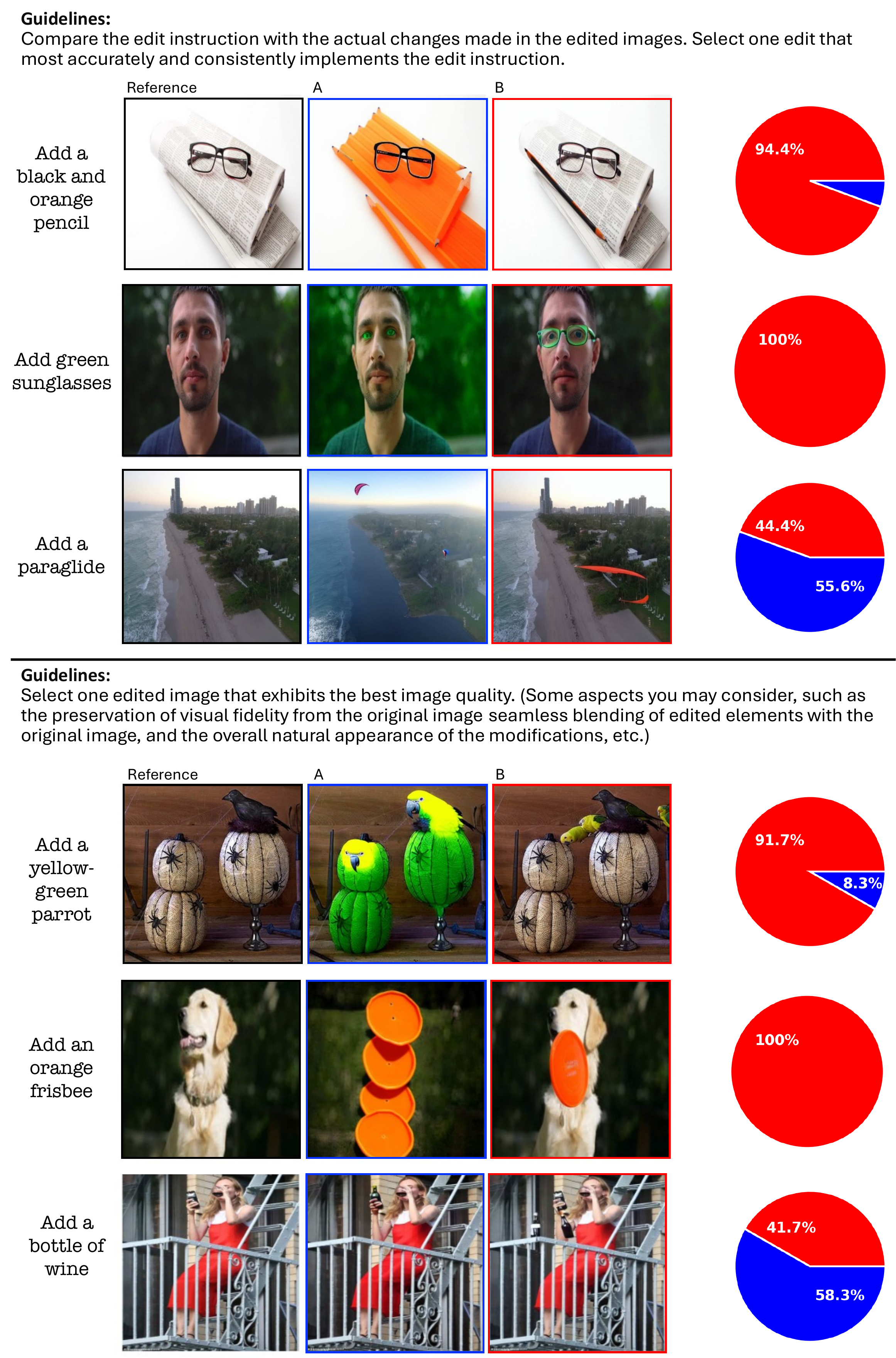}
    \caption{\textbf{Human Evaluation Examples}. Examples of the qualitative survey against IP2P alongside the response distribution (our method in \textcolor{red}{red} and the baseline in \textcolor{blue}{blue}).
    The examples include both successful and failed cases of our model.
    The first three top examples correspond with a question focused on the edit completion, and the three bottom ones on the resulting image quality. }
    \label{fig:human-survey}
\end{figure*}

While quantitative metrics are important for evaluating image editing performance, they do not fully capture human satisfaction with the edited outcomes.
To this end, we conduct a human evaluation survey, as explained in \Cref{sec:res2}, comparing our model with IP2P and hive (\cref{SM_table:human_hive}).
Following \citep{zhang2024magicbrush}, we pose two questions: one regarding the execution of the requested edit and another concerning the overall quality of the resulting images.
\Cref{fig:human-survey} illustrates examples from our human survey along with the questions posed.
Overall, our method leads to better results for human perception.
Interestingly, as expected due to how \pipe{} was constructed, our model maintains a higher level of consistency with the original images in both its success and failure cases.
For example, in the third row of \Cref{fig:human-survey}, while IP2P generates a more reliable paraglide, it fails to preserve the original background.

\begin{table}[t]
\centering
    \setlength\tabcolsep{0pt}
    \begin{tabular*}{1\linewidth}{@{\extracolsep{\fill}}lcccc} 
    \toprule
    \multirow{3}{*}{\textbf{Methods}} & \multicolumn{2}{c}{\textbf{Edit faithfulness}} & \multicolumn{2}{c}{\textbf{Quality}} \\
    & Overall & Per & Overall & Per- \\
    & [\%] & image & [\%] & image \\
    \midrule
    Hive  & 25.9 & 21 & 24.8 & 22 \\ 
    Ours & \textbf{74.1} & \textbf{79} & \textbf{75.2} & \textbf{78} \\
    \bottomrule 
    \end{tabular*}
    \captionsetup{skip=2pt} 
    \caption{\textbf{Human Evaluation against Hive. }}
    \label{SM_table:human_hive}
    \vspace{-0.4cm}
\end{table}

\section{Social Impact and Ethical Consideration}
\label{supp:social_impact}
Using \pipe{} or the model trained with it significantly enhances the ability to add objects to images based on textual instructions.
This offers considerable benefits, enabling users to seamlessly and quickly incorporate objects into images, thereby eliminating the need for specialized skills or expensive tools.
The field of image editing, specifically the addition of objects, presents potential risks.
It could be exploited by malicious individuals to create deceptive or harmful imagery, thus facilitating misinformation or adverse effects.
Users are, therefore, encouraged to use our findings responsibly and ethically, ensuring that their applications are secure and constructive.
Furthermore, \pipe{}, was developed using a VLM \citep{wang2023cogvlm} and an LLM \citep{jiang2023mistral}, with the model training starting from a SD checkpoint \citep{rombach2022high}.
Given that the models were trained on potentially biased or explicit, unfiltered data, the resulting dataset may reflect these original biases.

\end{document}